\documentclass[conference]{IEEEtran}
\usepackage{times}

\usepackage[numbers, sort&compress]{natbib}
\usepackage{multicol}
\usepackage{multirow}
\usepackage{graphicx}
\usepackage{booktabs}
\usepackage{mathtools}
\usepackage{subcaption}
\usepackage{amsfonts, amsmath}
\usepackage[bookmarks=true,hidelinks]{hyperref}

\def\dq{\dot{q}}
\def\ddq{\ddot{q}}

\begin{document}

\title{Data-driven Interpretable Hybrid Robot Dynamics}

\author{
Christopher E. Mower$^{1}$, Rui Zong$^{1}$, Haitham Bou-Ammar$^{1,2}$\\
$^{1}$Huawei, Noah's Ark Lab, London, UK\\
$^{2}$University College London, London, UK
}



%

\maketitle

\begin{abstract}
We study data-driven identification of interpretable hybrid robot dynamics, where an analytical rigid-body dynamics model is complemented by a learned residual torque term. Using symbolic regression and sparse identification of nonlinear dynamics (SINDy), we recover compact closed-form expressions for this residual from joint-space data. In simulation on a 7-DoF Franka arm with known dynamics, these interpretable models accurately recover inertial, Coriolis, gravity, and viscous effects with very small relative error and outperform neural-network baselines in both accuracy and generalization. On real data from a 7-DoF WAM arm, symbolic-regression residuals generalize substantially better than SINDy and neural networks, which tend to overfit, and suggest candidate new closed-form formulations that extend the nominal dynamics model for this robot. Overall, the results indicate that interpretable residual dynamics models provide compact, accurate, and physically meaningful alternatives to black-box function approximators for torque prediction.

\end{abstract}

\IEEEpeerreviewmaketitle

\section{Introduction}

Accurate estimation of a robot’s dynamics (i.e., its equations of motion) is crucial for applications such as force control~\cite{Iskandar23} and external torque estimation~\cite{Liu21}.
However, these dynamics depend on many interacting factors (e.g., mechanical structure, payload, contact conditions), making modeling difficult.
As a result, purely model-based approaches often become impractical, requiring considerable expertise to derive the models.

In parallel, the emergence of large-scale vision--language--action (VLA) models~\cite{rt2,openxembodiment,openvla} is rapidly changing how we think about robot controllers: instead of hand-designed pipelines, robots are increasingly treated as black-box embodiments that map images and language instructions to low-level actions.
However, current VLA datasets and policies are almost exclusively built around visual, linguistic, and proprioceptive signals; explicit force or torque measurements are rarely available at scale, and only a small number of recent works explore force- or touch-augmented foundation models~\cite{xie25forceful,tactilevla,forcevla}.
These works suggest that explicitly treating force and touch as first-class modalities can substantially improve performance on contact-rich manipulation tasks, but they also highlight a key bottleneck: high-quality force/torque and tactile sensors are expensive, difficult to integrate across many platforms, and often produce noisy, robot-specific signals, so most robots in large datasets simply do not have them.
If we could reliably infer internal and external joint torques from readily available signals (e.g.\ encoder-based kinematics and motor torques) using an interpretable dynamics model, we would effectively equip fleets of otherwise ``blind-to-force'' robots with virtual force sensors at negligible hardware cost.
This would both enable scalable collection of contact-rich supervision for future VLA-style policies and provide physically grounded signals (estimated forces and torques) that can be used as additional targets or auxiliary tasks during training, tightening the connection between high-level foundation models and the underlying mechanics of the robot.

From a reinforcement learning (RL) perspective, good physical priors over the robot’s dynamics are equally important. A long line of work in legged locomotion has shown that combining deep RL with accurate simulators and carefully designed structure can produce remarkably robust controllers for quadrupeds in the real world~\cite{Fankhauser13,Lee20,Rudin22,Ha25}. In these systems, the policy is typically trained in a high-fidelity rigid-body simulator with hand-engineered rewards and contact models, and then transferred to hardware via domain randomization and other robustness techniques. While this pipeline has enabled blind locomotion over challenging terrain~\cite{Lee20} and rapid policy training in minutes~\cite{Rudin22}, it remains heavily reliant on accurate but opaque simulators, and the learned policies themselves provide little direct insight into the underlying mechanics. Interpretable models of joint torques and contact forces offer a complementary route: they can serve as structured priors or residual models inside model-based or hybrid RL schemes, provide physically meaningful features and auxiliary targets for value and policy networks, and constrain exploration to behavior that remains consistent with known dynamics. In the context of contact-rich tasks surveyed by Ha et al.~\cite{Ha25}, such physically grounded priors are crucial for improving sample efficiency, safety, and out-of-distribution generalization, especially when training directly on hardware or across diverse robot morphologies.

Together with the rapid growth of AI techniques, these challenges have motivated researchers to explore data-driven approaches based on neural networks.
Although such models can outperform traditional model-based methods and adapt to new scenarios, their black-box nature raises concerns about trust and robustness.
In contrast, methods that derives explicit mathematical expressions directly from data, without assuming a predefined model structure, e.g,, symbolic/sparse regression.
This makes these methods particularly useful for describing physical phenomena that lack explicit mathematical formulations or are too complex to derive analytically.
Moreover, they have been observed to uncover new, interpretable relationships hidden in the data, 
making it well suited for approximating complex behaviors such as friction torque, whose precise characterization is difficult due to its dependence on multiple interacting factors.

A useful historical analogy is the progression from Kepler’s and Newton’s descriptions of planetary motion.
Kepler’s laws provided an accurate, data-driven description of orbits, but they remained essentially empirical and system-specific.
Newton’s formulation, on the other hand, introduced a simple and interpretable law of gravitation that explained Kepler’s observations, generalized far beyond the original data, and ultimately enabled engineered capabilities such as interplanetary navigation.
In a similar spirit, a highly accurate but opaque neural network model of joint torques plays a role akin to Kepler’s descriptive laws, whereas an explicit, low-complexity model of the robot dynamics corresponds more closely to Newton’s formulation: it can be inspected, trusted, and reused for analysis, controller design, and extrapolation beyond the training conditions.
This motivates the search for data-driven models of robot dynamics that are not only accurate, but also sparse and interpretable.

In this paper, we focus on developing interpretable and accurate dynamic models.
Concretely, we model joint-space dynamics as an analytical rigid-body model plus a data-driven residual, and investigate the affect of using symbolic regression and SINDy-style sparse regression on joint-space features to identify closed-form expressions for this residual from data.
We validate the approach in simulation on a 7-DoF Franka arm with known dynamics, showing that the recovered expressions accurately reconstruct known dynamics equations and substantially outperform neural-network baselines in both accuracy and generalization.
We then apply the same methodology to real data from a 7-DoF WAM arm, comparing symbolic and sparse models against neural networks, and analyzing how each class trades off in-sample fit, out-of-sample generalization, and interpretability.
Throughout, we emphasize models that not only predict joint torques well but also expose physically meaningful structure that can be inspected and potentially reused for tasks such as virtual force sensing, controller design, and physically grounded robot learning.

\section{Problem Formulation}

For an $n$-DoF robot, the joint-space dynamics is given by
\begin{equation}\label{eq:joint-space-dynamics}
    \tau_m = \tau_{dyn} + \tau_{ext},
\end{equation}
where 
$\tau_m \in \mathbb{R}^n$ are the motor joint torques, 
$\tau_{dyn} \in \mathbb{R}^n$ are the torques due to the internal robot dynamics, and
$\tau_{ext} \in \mathbb{R}^n$ are external joint torques.

A common model used to estimate $\tau_{dyn}$ is the rigid-body dynamics equations of motion
\begin{equation}
    \label{eq:rbd}
    \tau_{rbd}(q,\dot{q},\ddot{q}) 
    = M(q)\ddot{q} + C(q,\dot{q})\dot{q} + \tau_g(q),
\end{equation}
where 
$M(\cdot) \in \mathbb{R}^{n\times n}$ is the inertia matrix,
$C(\cdot) \in \mathbb{R}^{n\times n}$ is the Coriolis matrix,
$\tau_g(\cdot) \in \mathbb{R}^n$ is the gravitational torque, and
$q,\dot{q},\ddot{q} \in \mathbb{R}^n$ are the joint position, velocity, and acceleration, respectively.
For sufficiently rigid/heavy robots whose dynamics are dominated by inertial effects, 
$\tau_{rbd}$ often provides a good approximation of $\tau_{dyn}$.
In general, however, unmodeled effects (e.g., friction, actuator dynamics, or flexibilities) can lead to a significant mismatch between the true internal dynamics $\tau_{dyn}$ and the model $\tau_{rbd}$, i.e.
\begin{equation}
    \label{eq:dyn-err}
    \tau_{dyn} = \tau_{rbd} + \epsilon,
\end{equation}
where $\epsilon \in \mathbb{R}^n$ denotes the modeling error such that $\|\epsilon\|$ is non-negligible.
The error can depend on nonlinear relationship with the motion of the joints, coupling effects of the joints, manufacturing mismatches, and potentially many other effects (e.g., temperature).
Deriving a model for $\epsilon$ by hand is error-prone and can be tedious for researchers. 
On the other hand, as mentioned above, neural networks have been observed to be good predictors of $\epsilon$, however, they are not interpretable, leading to lack of trust in real world applications.
The goal of this work is to discover an interpretable, data-driven model for the error term $\epsilon$ in \eqref{eq:dyn-err}.

\section{Related Work}

Before studying residual dynamic models, obtaining an accurate rigid-body model is essential. As early as 1986, Atkeson~\cite{atkeson} demonstrated that inertial parameters can be identified using linear regression and least-squares techniques, showing on the PUMA manipulator that parameters estimated from the regressor form $w = A\,\phi$ produce torque predictions that significantly outperform CAD-based models. Sousa~\cite{sousa2014} later introduced the notion of enforcing physical feasibility in inertial parameter identification, observing that the physically feasible inertial parameters form a convex set and that these constraints can be expressed as Linear Matrix Inequalities (LMIs). This allows the identification problem to be solved via semidefinite programming while guaranteeing that $\phi$ corresponds to a reliable mass distribution. Their experiments on the WAM manipulator demonstrated a framework for constrained and physically feasible parameter identification. 
The optimization problem required to identify these dynamic parameters is often ill-conditioned; consequently, a complementary line of work has focused on generating optimal excitation trajectories~\cite{Swevers97, tian2024excitation}.

While parameter identification for robotic manipulators is now relatively mature, the same process becomes substantially more challenging for legged robots because the common assumption of zero external torques does not hold. Khorshidi~\cite{Khorshidi25} addresses this difficulty by projecting the robot dynamics into the null space of contact forces with the null-space projector $  P(q) = I_m - J_c(q)^\dagger J_c(q)$, yielding torque equations that are independent of ground reaction forces. They then identify the inertial parameters by solving the constrained optimization problem $\hat{\phi} = \arg\min_{\phi}\|Y\phi - \tau_m\|^2 \ \text{s.t.}\ \phi \in \mathcal{P}_{\mathrm{phys}}$, where $\mathcal{P}_{\mathrm{phys}}$ encodes mass positivity, center-of-mass bounds, and inertia LMIs. The resulting physically consistent model achieves more reliable joints torque predictions than neural-network-based estimators, maintaining low prediction error across a wide range of tasks.

A widely adopted strategy for improving robot dynamics models is to learn the residual mismatch between the true dynamics and the nominal model, because this residual is usually much smoother than the full dynamics and can thus be captured effectively with Gaussian Processes (GPs). Carron~\cite{Carron} implement this idea by training a GP prior to approximate the model error $d(x,u)$ in the discrete-time system dynamics $x_{k+1} = A x_k + B u_k + B_d\!\left(d(x_k,u_k) + \bar d\right)$, where the constant offset noise $\bar d$ is estimated online via an extended Kalman Filter (EFK). In their formulation, the GP mean $\mu_d(x,u)$ provides a feedforward compensation to the nominal inverse-dynamics model, while the GP covariance $\Sigma_d(x,u)$ contributes to the predicted state covariance $\Sigma_x$, which is used in chance constraints within the Model Predictive Control (MPC) horizon. Experiments on a compliant 6-DoF robotic arm show substantial tracking improvements compared to both PID and nominal MPC baselines. However, given its intrinsic properties, GP-based residual modeling is fundamentally limited. Because the GP is trained offline and must be sparsified to remain computationally feasible at 1 kHz control rates, this residual model cannot adapt to changing dynamics. Moreover, GP models cannot automatically capture complex physical effects without appropriately designed kernels, which leads to a strong sensitivity to hyperparameter tuning. Its accuracy is constrained by the risk of overfitting, which ultimately highlights the strong dependence of GP-based residual models on the distribution of the training task data.

Recently, Scholl used symbolic regression to learn an interpretable friction model for the KUKA LWR robot arm~\cite{Scholl25} where $\epsilon^{(j)} = \tau_f^{(j)}(\dq^{(j)}, \tau_g^{(j)})$ and $j$ indicates the model is per joint. 
By operating the robot at low, near-constant velocities, i.e. $\dq,\ddq,C(q, \dq)\dq \approx 0_n$, and assuming zero external torques $\tau_{ext} =0_n$ they were able to simplify the dynamic model $\tau_m = \tau_g(q) + \epsilon$ and thus uncover a model for $\epsilon$ using symbolic regression.
Whilst the work by Scholl et al provides early validation that symbolic regression is useful for modeling dynamic effects, the work is limited. 
Their approach identifies a quasi-static, per-joint friction law only in a narrow operating regime: near-zero accelerations, low velocities, and single-joint motions. 
As a result, the learned model does not account for dynamic friction phenomena (e.g. hysteresis, pre-sliding) or joint–joint coupling effects, and it strongly relies on an accurate rigid-body model to remove inertial and Coriolis contributions. 
Moreover, the symbolic expressions are unconstrained outside the identification domain, so there is no guarantee that the inferred friction model extrapolates sensibly to higher velocities, larger accelerations, or more aggressive, multi-joint trajectories.

\section{Methods for Interpretable Dynamics}

Two common approaches are considered in this work for inferring interpretable dynamics models from motion data.

\subsection{Symbolic regression}
\label{sec:sr}

Symbolic regression is a supervised learning approach in which the goal is to discover an explicit analytic expression that relates input variables to a target quantity, rather than fitting the parameters of a fixed model class.  Given a dataset
$\mathcal{D} = \{(x_i, y_i)\}_{i=1}^N$,
symbolic regression searches over a space $\mathcal{F}$ of candidate expressions built from basic operators (e.g.\ $+,-,\times,/,\exp,\log,\sin,\dots$), input variables, and free numerical constants to find a closed-form function
$f : \mathbb{R}^d \to \mathbb{R}$
that explains the data~\cite{pysr}.

Formally, one can view symbolic regression as the problem
\begin{equation}
\widehat{f} = \underset{f \in \mathcal{F}}{\text{arg}\min}~\mathcal{L}\bigl(f; \mathcal{D}\bigr),
\label{eq:symbolic-regression-objective}
\end{equation}
where $\mathcal{L}$ is a loss function that evaluates candidate expressions $f$ on the data.  
In contrast to typical regression techniques, where the functional form is fixed in advance (e.g.\ linear models, polynomials of fixed degree, neural networks) and only the parameters are optimized, symbolic regression treats both the structure of the expression and its numerical constants as unknowns.  
Thus, the loss $\mathcal{L}$ is typically designed to reflect a trade-off between predictive accuracy and some notion of simplicity, so that the selected expressions are not only accurate on the observed data but also compact and interpretable.  
This is beneficial compared to models such as neural networks, which are often treated as black boxes, since the resulting expressions can be directly inspected and analyzed.  
Modern tools such as PySR~\cite{pysr} implement this paradigm using heuristic search procedures to explore the large, discrete space $\mathcal{F}$ while maintaining this balance between fit quality and model simplicity, and have been used to derive physics models in various applications (e.g.\ particle physics~\cite{MoralesAlvarado24}).

\subsection{Sparse identification of Nonlinear Dynamics}
\label{sec:sindy}

Sparse identification of nonlinear dynamics (SINDy) is a data-driven framework for discovering governing equations of dynamical systems from time-series data~\cite{sindy}.  
Given measurements of the system state $\{x_i\}_{i=1}^N$ and corresponding (estimated) time derivatives $\{\dot{x}_i\}_{i=1}^N$, SINDy assumes that the dynamics can be expressed as a sparse linear combination of candidate nonlinear functions.  
To this end, one constructs a feature library
$\Theta(X) \in \mathbb{R}^{N \times p}$,
whose columns contain nonlinear transformations of the states (and possibly inputs), such as polynomials, trigonometric functions, or other user-specified basis functions.  
The dynamical model is then written as
$\dot{X} \approx \Theta(X)\,\Xi$
where $\dot{X} \in \mathbb{R}^{N \times d}$ stacks the time derivatives and $\Xi \in \mathbb{R}^{p \times d}$ is a matrix of coefficients.  
Identifying the dynamics reduces to solving a regression problem for $\Xi$ with a sparsity-promoting procedure (e.g.\ sequentially thresholded least squares), so that only a small subset of candidate terms remains active in each column.  
The resulting models are compact and interpretable, since each retained term corresponds to a specific mechanism in the dynamics.

In this work, we do not use SINDy to identify an explicit evolution law $\dot{x} = f(x)$, but instead adapt the same sparse-regression machinery to learn a general input--output mapping $y = g(x)$.  
Conceptually, this amounts to replacing the time-derivative matrix $\dot{X}$ in the standard SINDy formulation by a target matrix $Y$ collecting the quantities of interest, while keeping the construction of the nonlinear feature library and the sparsity-enforcing regression procedure unchanged.  
The identification step then becomes
\begin{equation}
  Y \approx \Theta(X)\,\Xi,
\end{equation}
so that one recovers a sparse combination of basis functions that best predicts the target signal from the input features.  
This viewpoint casts our approach as a SINDy-inspired sparse polynomial regression, reusing the standard SINDy components (feature libraries, sparsity-promoting solvers, and their implementation in PySINDy~\cite{pysindy}) in a supervised regression setting rather than for explicit time-derivative modeling.

\section{Experiments}

In this section, we report our experimental findings.

\subsection{Validation of pipeline in simulation}

\begin{figure}
    \centering
    \includegraphics[width=\linewidth]{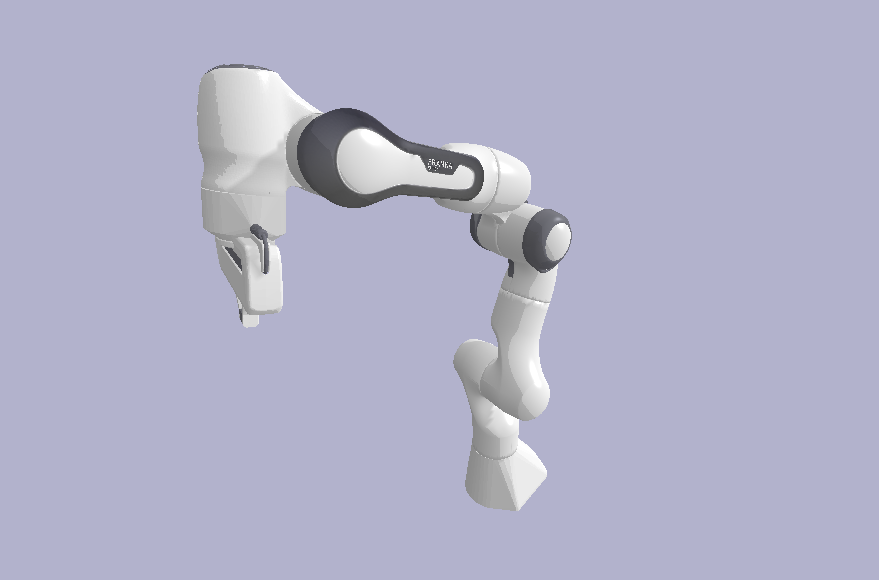}
    \caption{Visualization of the simulated 7-DoF Franka Robot arm. Note, in our experiments we only use PyBullet to visualize the robot, we use our own simulator definition.}
    \label{fig:franka-sim}
    \vspace{-0.5cm}
\end{figure}

A simulator was developed using a fixed-base 7-DoF Franka Emika Panda robot arm, shown in Fig.~\ref{fig:franka-sim}. 
In these initial experiments, our primary objective is to validate the overall pipeline and evaluation infrastructure for the approaches considered in this study. 
It is important to note that PyBullet~\cite{pybullet} is used only as a visualizer in this initial experimental setup;
all robot dynamics are simulated using a known analytical rigid-body model of the Franka arm implemented in Pinocchio~\cite{pinocchio}.
This design choice provides a clean benchmark in which modeling errors and low-level simulation artifacts are removed, allowing us to attribute the observed behavior directly to the algorithms under study.
Note, the simulator developed in this section assumes no external torques, i.e, $\tau_{ext} = 0_7$.
This means by~\eqref{eq:joint-space-dynamics}, that predicting $\tau_{dyn}$ is equivalent to predicting $\tau_m$.

\subsubsection{Simulator}

The robot is modeled as a 7-DoF manipulator with joint configuration
$q \in \mathbb{R}^7$ and joint velocities $\dot{q} \in \mathbb{R}^7$.
The joint-space dynamics are written as
\begin{equation}\label{eq:franka-dynamics}
  M(q)\ddot{q} + C(q,\dot{q})\dot{q} + \tau_g(q) = \tau_{eff},
\end{equation}
where $M(q) \in \mathbb{R}^{7 \times 7}$ is the joint-space inertia matrix,
$C(q,\dot{q})\dot{q}$ collects Coriolis and centrifugal terms, and
$\tau_g(q)$ is the generalized gravity torque.

A diagonal viscous damping model is used,
\begin{equation}\label{eq:franka-damping}
  \tau_{d}(\dot{q}) = D\dot{q},
\end{equation}
where $D = \text{diag}(d_1, \dots, d_7) \in \mathbb{R}^{7 \times 7}$ is a diagonal matrix with positive diagonal elements, i.e., $d_i > 0$ for all $i=1{:}7$.
Given commanded motor torques $\tau_m \in \mathbb{R}^7$, the effective torques \eqref{eq:franka-dynamics} are found by the residual
\begin{equation}\label{eq:franka-tau-eff}
  \tau_{eff} = \tau_m - \tau_{d}.
\end{equation}
where $\tau_d = \tau_d(\dot{q})$.
By including the damping model~\eqref{eq:franka-damping} in~\eqref{eq:franka-tau-eff}, we introduce a known, structured contribution to the effective torque that can serve as additional signal for the considered methods to identify and recover from data.

The joint accelerations are computed by solving~\eqref{eq:franka-dynamics} for $\ddot{q}$, forming the forward dynamics, given by
\begin{equation}\label{eq:franka-aba}
  \ddot{q}
  = M(q)^{-1}\bigl(\tau_{eff} - C(q,\dot{q})\,\dot{q} - \tau_g(q)\bigr).
\end{equation}
We use Pinocchio's efficient implementation of the articulated-body algorithm to compute~\eqref{eq:franka-aba}.

Given the state $(q_s, \dot{q}_s)$ and the effective torque $\tau_{eff,s}$ at time step $s$, 
we first compute $\ddot{q}_s$ using~\eqref{eq:franka-aba} and then the state is integrated using a semi-implicit
(symplectic) Euler scheme with simulation time step $\Delta t_{sim}$.
Thus, the next state is computed using
\begin{equation}\label{eq:franka-simulator-integration}
\begin{aligned}
  \dot{q}_{s+1} &= \dot{q}_s + \Delta t_{sim}\,\ddot{q}_s, \\
  q_{s+1}      &= q_s + \Delta t_{sim}\,\dot{q}_{s+1}.
\end{aligned}
\end{equation}

This yields a deterministic simulator with known dynamics and full access to the
true joint state.
By substituting~\eqref{eq:franka-tau-eff} into~\eqref{eq:franka-dynamics} and rearranging for $\tau_m$, 
provides us the model that we hope to uncover from motion data.
The model we hope to discover is thus given by
\begin{equation}\label{eq:franka-model}
    \tau_m = \tau_i + \tau_c + \tau_g + \tau_{d}
\end{equation}
where $\tau_i = M(q)\ddot{q}$, $\tau_c = C(q,\dot{q})\dot{q}$, and $\tau_g = \tau_g(q)$.

\subsubsection{Controller}

The low-level controller, used to compute the motor torques $\tau_m$ is a joint-space PID law with gravity compensation. 
Given the desired joint positions
$q^\star \in \mathbb{R}^7$ and velocities $\dot{q}^\star \in \mathbb{R}^7$,
the motor torques are given by
\begin{equation}\label{eq:franka-controller}
\begin{aligned}
  \tau_m
    &= \tau_g(q) + \tau_{pid},\\
  \tau_{pid}
    &= K_p \bigl(q^\star - q\bigr)
     + K_i \int (q^\star - q)\,dt
     + K_d \bigl(\dot{q}^\star - \dot{q}\bigr)
\end{aligned}
\end{equation}

where $\tau_g(q)$ is the generalized gravity torque computed by Pinocchio,
and $K_p,K_i,K_d\in\mathbb{R}^7$ are diagonal matrices representing the PID gains.
The integral term is updated at the controller (environment) sampling period $\Delta t_{env}$
and is clipped component-wise to prevent windup.

This setup provides a clean testbed in which both the plant
and the controller share the same perfect model of the robot, isolating the
behavior of the proposed methods from modeling errors and sensor noise.

\subsubsection{Data generation}

In order to test our proposed data-driven approach, we require data collected from the simulator.
This section describes the data generation and simulation steps.

\paragraph{Trajectory generation}

To excite the dynamics of all joints, we generate joint-space reference trajectories as sums of sinusoidal signals.
For each rollout, we first sample a random initial configuration $q_0 \in \mathbb{R}^7$ inside the joint limits of the Franka arm, with a fixed margin subtracted from each bound to avoid saturating the joints.
Formally, if $q_{\min}, q_{\max} \in \mathbb{R}^7$ denote the lower and upper joint limits respectively and $0 < \rho \in \mathbb{R}$ is a fixed margin, we sample
\begin{equation}
  q_{0} \sim \mathcal{U}\bigl(q_{\min} + \rho e,\, q_{\max} - \rho e\bigr),
\end{equation}
where $\mathcal{U}(a, b)$ represents a uniform distribution with lower and upper bounds $a,b \in \mathbb{R}^7$ such that $a_j < b_j$ for all $j=1{:}7$, and $e=[1, \dots, 1]^T\in\mathbb{R}^7$ is the vector of ones.
We then construct a time grid $\{t_k\}_{k=0}^{N}$ over a horizon $T$ with sampling period $\Delta t_{env}$.
For each joint $j=1{:}7$, we form a multi-sine signal
\begin{equation}
  q_{raw,j}(t) = \sum_{\ell=1}^{n_m} a_{j\ell} \sin\bigl(2\pi f_{j\ell} t + \phi_{j\ell}\bigr),
\end{equation}
where $n_m$ is the number of sinusoidal modes per joint and the amplitudes $a_{j\ell}$, frequencies $f_{j\ell}$, and phases $\phi_{j\ell}$ are drawn independently from uniform distributions over prescribed ranges.
The corresponding velocity signal is obtained analytically as
\begin{equation}
  \dot{q}_{raw,j}(t) = \sum_{\ell=1}^{n_m} a_{j\ell} (2\pi f_{j\ell}) \cos\bigl(2\pi f_{j\ell} t + \phi_{j\ell}\bigr).
\end{equation}

Each pair $\bigl(q_{raw,j}, \dot{q}_{raw,j}\bigr)$ is then scaled by a single factor such that the resulting joint positions and velocities remain within the (shrunken) position limits and the nominal velocity limits of joint $j$.
Formally, for each joint $j$ we define the available position margin around the initial configuration $q_{0,j}$ as
\begin{equation}
  r_j = \min\bigl(q_{\max,j} - q_{0,j},\; q_{0,j} - q_{\min,j}\bigr) - \rho,
\end{equation}
where $q_{\min,j}$ and $q_{\max,j}$ denote the $j$th components of $q_{\min}$ and $q_{\max}$, respectively.
Let $\dot{q}_{\max} \in \mathbb{R}^7$ denote the joint-velocity limits and define
\begin{align}
  \alpha_{pos,j} &= \frac{r_j}{\max_k \bigl|q_{raw,j}(t_k)\bigr| + \varepsilon}, \\
  \alpha_{vel,j} &= \frac{\dot{q}_{\max,j}}{\max_k \bigl|\dot{q}_{raw,j}(t_k)\bigr| + \varepsilon},
\end{align}
with a small $0<\varepsilon \in\mathbb{R}$ such that $|\varepsilon|\ll 1$ to avoid division by zero.
The final scaling factor is then
\begin{equation}
  s_j = \min\bigl(\alpha_{pos,j},\; \alpha_{vel,j}\bigr),
\end{equation}
and the discrete desired trajectories are obtained as
\begin{align}
  q^\star_j(t_k)       &= q_{0,j} + s_j\,q_{raw,j}(t_k), \\
  \dot{q}^\star_j(t_k) &= s_j\,\dot{q}_{raw,j}(t_k).
\end{align}
This yields discrete desired trajectories $q^\star(t_k)$ and $\dot{q}^\star(t_k)$.

Using the above procedure and 

\paragraph{Simulation}

Given desired trajectories $q^\star(t_k), \dot{q}^\star(t_k)$ for $k=0{:}N$, we simulate the closed-loop system consisting of the Franka simulator and the joint-space PID controller described in the previous sections.
The initial state is set to
\begin{equation}
  q_0 = q^\star(t_0), \qquad \dot{q}_0 = \dot{q}^\star(t_0).
\end{equation}
At each environment step $k = 0{:}N-1$, the controller computes a motor torque command $\tau_m$ using~\eqref{eq:franka-controller}, which is held constant while the plant is integrated for a fixed number of smaller integration sub-steps using the semi-implicit Euler scheme~\eqref{eq:franka-simulator-integration}.
The resulting joint states $q_{k+1}$ and $\dot{q}_{k+1}$ are recorded at the environment rate $\Delta t_{env}$.

For each rollout we thus obtain sequences $\{q_k\}_{k=0}^{N}, \{\dot{q}_k\}_{k=0}^{N}, \{\tau_{m,k}\}_{k=0}^{N-1}$,
representing the joint positions, joint velocities, and motor torque commands over the horizon, respectively, together with the reference signals $q^\star(t_k)$, $\dot{q}^\star(t_k)$ and the time stamps $\{t_k\}_{k=0}^{N}$.
Each rollout is stored on disk as a separate trajectory, and a dataset is obtained by repeating this procedure for multiple independently sampled reference trajectories.

\begin{figure*}[ht]
    \centering
    \includegraphics[width=\linewidth]{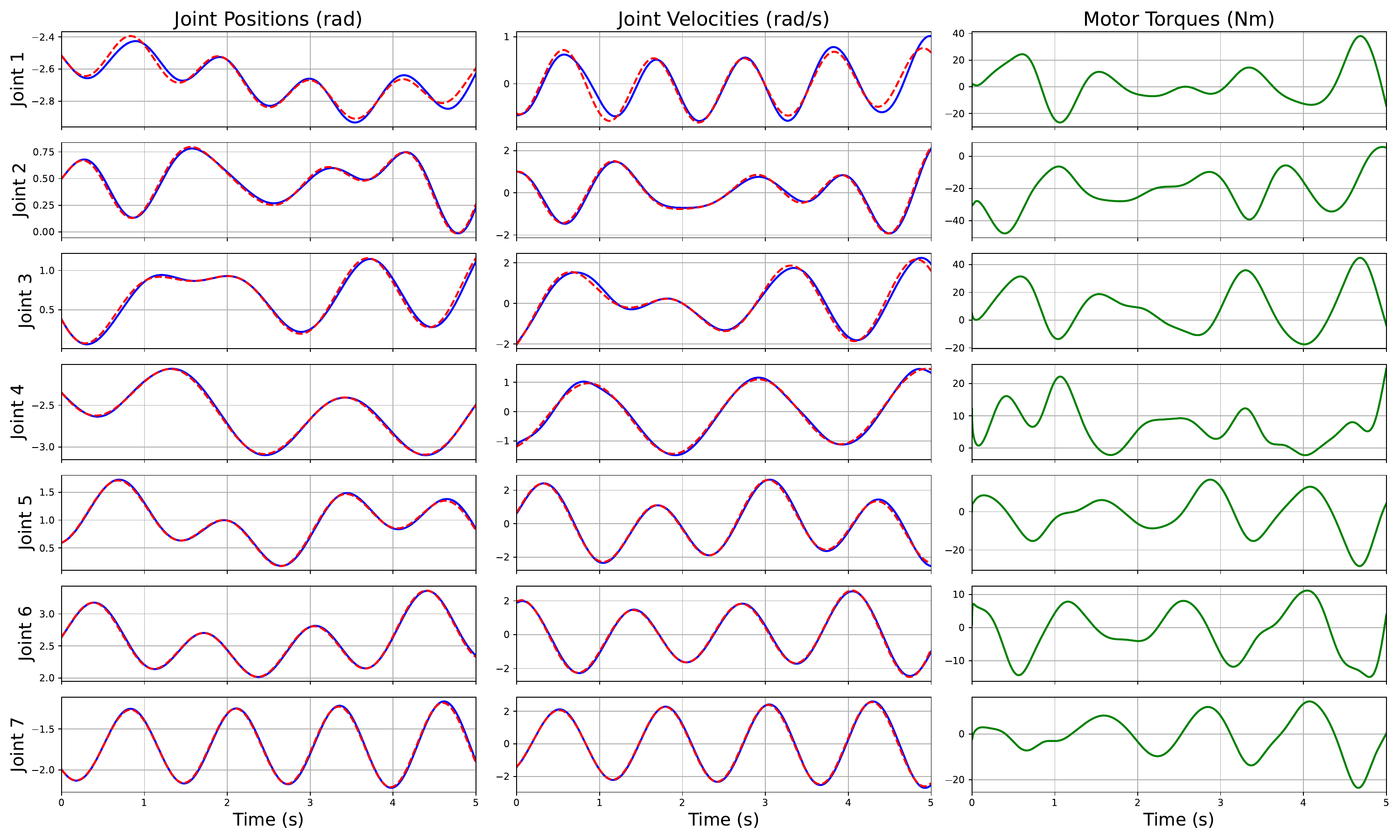}
    \caption{Example of a trajectory collected from our idealized simulator based on the 7-DoF Franka robot arm used in our validation experiments. Motor torques were computed using a controller implementing PID with gravity compensation.}
    \label{fig:franka-traj}
    \vspace{-0.5cm}
\end{figure*}

We used the above procedure to generate 10 training trajectories and 10 test trajectories. 
An example of a trajectory generated by the above procedure and resulting rollout in the simulator are shown in Figure~\ref{fig:franka-traj}.

\subsubsection{Training and evaluation}

The trajectories collected are split into disjoint training and test sets.
Each candidate data-driven dynamics approach takes as input the features 
$q, \dot{q}, \ddot{q}, \dddot{q}, \tau_i, \tau_c, \tau_g$ (i.e, the joint position, velocity, acceleration, jerk, inertia torques, Coriolis/centrifugal torques, and gravity torques respectively),
and learns a mapping, such that $f:\mathbb{R}^{49}\rightarrow\mathbb{R}^7$, which estimates the motor torques $\tau_m$,
\begin{equation}
  \widehat{\tau}_m = f(q, \dot{q}, \ddot{q}, \dddot{q}, \tau_i, \tau_c, \tau_g).
\end{equation}
Note, the joint accelerations $\ddot{q}$ and jerk $\dddot{q}$ are estimated using finite-differencing method.

The models $f$ are fit on the training set only, and after training, we evaluate each model on both the training and test sets.
For each set, we reconstruct the feature matrices and compute the corresponding torque predictions $\widehat{\tau}_m$.
Performance is quantified by a relative root-mean-square error (RMSE) per joint across each dataset.

\subsubsection{Methods compared}
\label{sec:methods-compared}

As mentioned above, in the simulated setting, we have $\tau_{ext} = 0$ in
\eqref{eq:joint-space-dynamics}, so that $\tau_m = \tau_{dyn}$.
Using the decomposition in \eqref{eq:dyn-err}, each method either
approximates $\tau_{dyn}$ directly from data, or explicitly uses the
rigid-body model \eqref{eq:rbd} and learns an approximation of the
error term $\epsilon$.

All models take as input the same feature vector
\begin{equation}
  x
  =
  \begin{bmatrix}
    q^\top &
    \dot{q}^\top &
    \ddot{q}^\top &
    \dddot{q}^\top &
    \tau_i^\top &
    \tau_c^\top &
    \tau_g^\top
  \end{bmatrix}^\top
  \in \mathbb{R}^{49},
\end{equation}
constructed from the joint signals and the analytically computed
inertia, Coriolis/centrifugal, and gravity torques.
For the 7-DoF Franka arm used in the experiments, this corresponds to
$7$ variables of each type.
Given $x$, each model returns an estimate $\widehat{\tau}_m(x)$.
Symbolic-regression and SINDy-based models provide explicit closed-form
expressions, whereas neural-network models are treated as
black-box baselines.

\paragraph{Symbolic Regression (SR)}
The SR baseline directly learns a mapping
$\tau_{sr} : \mathbb{R}^{49} \rightarrow \mathbb{R}^7$
to approximate $\tau_{dyn}$ (and hence $\tau_m$) in
\eqref{eq:dyn-err}, without explicitly using \eqref{eq:rbd}.
We stack all training samples into matrices
$X\in\mathbb{R}^{N\times 49}$ and $Y\in\mathbb{R}^{N\times 7}$, where
each row of $X$ is a feature vector $x$ and each row of $Y$ is the
corresponding torque measurement $\tau_m$.
For each joint $j=1{:}7$ we fit an independent scalar model
\begin{equation}
  \tau_{sr,j}(x) \approx \tau_{m,j},
\end{equation}
using PySR~\cite{pysr}, with a search space of algebraic expressions built from
the input variables and the binary operators $\{+,-,*\}$ (division is
excluded).
PySR is run in batching mode with batch size $10^4$, and the internal
selection criterion balances mean-squared error on $Y$ and expression
complexity.
The result is a set of joint-wise, closed-form expressions
$\tau_{sr,j}$ that approximate $\tau_{dyn,j}$.

\paragraph{Sparse Identification of Nonlinear Dynamics (SINDy)}
The SINDy baseline, based on the work of Brunton et al~\cite{sindy}, also approximates $\tau_{dyn}$ directly, but
constrains the model to be a sparse multivariate polynomial in the
features.
Starting from $X$ and $Y$ as above, we construct a polynomial feature
library $\Theta(X)\in\mathbb{R}^{N\times p}$ using a basis of total
degree up to~2, including all pairwise interactions and a bias term.
We then learn a linear model
\begin{equation}
  \tau_{sindy}(x) = W\,\Theta(x) + b,
\end{equation}
with $W\in\mathbb{R}^{7\times p}$ and $b\in\mathbb{R}^7$ fitted
jointly for all seven outputs using sequentially thresholded
least-squares (STLSQ) with a sparsity threshold of $0.01$,
regularization parameter $\alpha = 10^{-4}$, and at most $100$
iterations.
The resulting $\tau_{sindy}$ is an explicit sparse polynomial
approximation of $\tau_{dyn}$ in \eqref{eq:dyn-err}.
To implement this approach we utilized the PySINDy library~\cite{pysindy}.

\paragraph{Hybrid RBD and SR (r-SR)}
The r-SR model uses the decomposition in \eqref{eq:dyn-err} explicitly.
For each training sample, we compute $\tau_{rbd}(q,\dot{q},\ddot{q})$
from the known Franka model and form the residual target
\begin{equation}
  y = \tau_m - \tau_{rbd}.
\end{equation}
We then train seven independent PySR models on $(x,y_j)$, exactly as
in the SR baseline, obtaining a residual estimate
\begin{equation}
  \widehat{\epsilon}(x)
  =
  \begin{bmatrix}
    \widehat{\epsilon}_1(x) &
    \dots &
    \widehat{\epsilon}_7(x)
  \end{bmatrix}^\top
  \in \mathbb{R}^7.
\end{equation}
At test time, the prediction is
\begin{equation}
  \widehat{\tau}_m^{\text{r-SR}}(x)
  = \tau_{rbd}(q,\dot{q},\ddot{q}) + \widehat{\epsilon}(x),
\end{equation}
so that r-SR can be interpreted as a symbolic model of the error term
$\epsilon$ in \eqref{eq:dyn-err}.

\paragraph{Hybrid RBD and SINDy (r-SINDy)}
The r-SINDy model follows the same residual-learning strategy, but
represents $\epsilon$ via a SINDy model.
After computing $\tau_{rbd}$, we form residual targets
\begin{equation}
  Y_{res} = \tau_m - \tau_{rbd}
\end{equation}
and reuse the same polynomial library $\Theta(X)$ as in the SINDy
baseline.
We then fit
\begin{equation}
  \widehat{\epsilon}(x) = W_{res}\,\Theta(x) + b_{res},
\end{equation}
using STLSQ with identical hyperparameters.
The hybrid prediction is
\begin{equation}
  \widehat{\tau}_m^{\text{r-SINDy}}(x)
  = \tau_{rbd}(q,\dot{q},\ddot{q}) + \widehat{\epsilon}(x),
\end{equation}
yielding an interpretable sparse polynomial approximation of the error
term in \eqref{eq:dyn-err}.

\paragraph{Hybrid RBD, SINDy, and SR (r-SINDy-SR)}
The r-SINDy-SR model refines the r-SINDy approximation of $\epsilon$
by adding a second, symbolic residual layer.
First, a hybrid RBD–SINDy model
\begin{equation}
  \widehat{\tau}_m^{\text{r-SINDy}}(x)
\end{equation}
is trained as above and kept fixed.
We then define a second residual
\begin{equation}
  y^{(2)} = \tau_m - \widehat{\tau}_m^{\text{r-SINDy}}(x)
\end{equation}
and train joint-wise PySR models on $(x,y^{(2)}_j)$, obtaining a
symbolic correction $\widehat{\epsilon}^{(2)}(x)$.
The final prediction is
\begin{equation}
  \widehat{\tau}_m^{\text{r-SINDy-SR}}(x)
  = \widehat{\tau}_m^{\text{r-SINDy}}(x) + \widehat{\epsilon}^{(2)}(x),
\end{equation}
which can be viewed as a two-layer approximation of the error term
$\epsilon$ in \eqref{eq:dyn-err}.

\paragraph{Neural Network (NN)}
The NN baseline directly parametrizes $\tau_{dyn}$ by a fully
connected neural network.
We use a multilayer perceptron
\begin{equation}
  \tau_{nn}(x) = f_{nn}(x),
\end{equation}
with two hidden layers of width $128$ and rectified-linear (ReLU)
activations.
The network is trained on the concatenated dataset using mean-squared
error loss between $\tau_{nn}(x)$ and $\tau_m$, the Adam optimizer
with learning rate $10^{-4}$, batch size $1024$, and $100$ training
epochs.
In the notation of \eqref{eq:dyn-err}, this model implicitly learns
$\tau_{dyn}$ (and hence $\epsilon$) without imposing an explicit
structure.

\paragraph{Hybrid RBD and NN (r-NN)}
Finally, the r-NN model combines the rigid-body model with a neural
approximation of the error term $\epsilon$ in \eqref{eq:dyn-err}.
As in the other hybrid methods, we first compute $\tau_{rbd}$ and form
residual targets
\begin{equation}
  y = \tau_m - \tau_{rbd}.
\end{equation}
We then train an ``error network'' $f_{err}$, with the same
architecture as the NN baseline, to map $x$ to an estimate
\begin{equation}
  \widehat{\epsilon}(x) = f_{err}(x).
\end{equation}
During training, the loss is evaluated on
\begin{equation}
  \tau_{rbd}(q,\dot{q},\ddot{q}) + \widehat{\epsilon}(x)
\end{equation}
against the measured $\tau_m$, so that the network explicitly learns
to correct the analytical model.
At test time the hybrid prediction is
\begin{equation}
  \widehat{\tau}_m^{\text{r-NN}}(x)
  = \tau_{rbd}(q,\dot{q},\ddot{q}) + \widehat{\epsilon}(x),
\end{equation}
which matches the decomposition in \eqref{eq:dyn-err} with
$\widehat{\epsilon}$ represented by a neural network.

\subsubsection{Results}

\begin{table}[t]
\centering
\small 
\setlength{\tabcolsep}{3pt} 
\caption{Training set RMSE per joint for the Franka simulation.}
\label{tab:franka-rmse_train}
\begin{tabular}{@{}cccccccc@{}}
\hline
Joint & SR & SINDy & r-SR & r-SINDy & r-SINDy-SR & NN & r-NN \\
\hline
1 & 0.002 & \textbf{0.002} & 0.002 & 0.002 & \textbf{0.002} & 0.072 & 0.041 \\
2 & \textbf{0.001} & \textbf{0.001} & \textbf{0.001} & \textbf{0.001} & \textbf{0.001} & 0.092 & 0.025 \\
3 & 0.002 & \textbf{0.002} & 0.002 & 0.002 & \textbf{0.002} & 0.084 & 0.062 \\
4 & 0.002 & 0.002 & 0.002 & 0.002 & \textbf{0.002} & 0.065 & 0.041 \\
5 & \textbf{0.002} & 0.003 & 0.003 & 0.003 & 0.003 & 0.138 & 0.071 \\
6 & \textbf{0.002} & 0.003 & 0.003 & 0.003 & 0.003 & 0.163 & 0.087 \\
7 & 0.003 & 0.003 & 0.003 & 0.003 & \textbf{0.003} & 0.168 & 0.111 \\
\hline
\end{tabular}

\caption{Test set RMSE per joint for the Franka simulation.}
\label{tab:franka-rmse_test}
\begin{tabular}{@{}cccccccc@{}}
\hline
Joint & SR & SINDy & r-SR & r-SINDy & r-SINDy-SR & NN & r-NN \\
\hline
1 & 0.002 & \textbf{0.002} & 0.002 & 0.002 & \textbf{0.002} & 0.687 & 0.994 \\
2 & 0.001 & 0.001 & 0.001 & 0.001 & \textbf{0.001} & 0.267 & 0.244 \\
3 & 0.002 & \textbf{0.002} & 0.002 & 0.002 & \textbf{0.002} & 0.388 & 0.628 \\
4 & 0.002 & 0.002 & 0.002 & 0.002 & \textbf{0.002} & 0.300 & 0.251 \\
5 & \textbf{0.003} & 0.003 & 0.003 & 0.003 & 0.003 & 0.679 & 0.624 \\
6 & \textbf{0.003} & 0.003 & 0.003 & 0.003 & 0.003 & 1.111 & 0.424 \\
7 & \textbf{0.004} & \textbf{0.004} & \textbf{0.004} & \textbf{0.004} & \textbf{0.004} & 1.356 & 0.986 \\
\hline
\end{tabular}
\vspace{-0.5cm}
\end{table}

Table~\ref{tab:franka-rmse_train} reports the per-joint relative RMSE
on the training set, while Table~\ref{tab:franka-rmse_test} shows the
corresponding errors on the test set.
Across both sets, the SR, SINDy, r-SR, r-SINDy, and r-SINDy-SR models
achieve virtually identical performance, with relative RMSEs on the
order of $10^{-3}$–$10^{-2}$ for all seven joints.
In particular, the best-performing method for each joint (shown in
bold) is always one of these interpretable models, and the numerical
differences between them are negligible at the scale of the metric.
This indicates that, in the simulated setting where
$\tau_{rbd}$ is computed from the same analytical model used to
generate the data, both the purely data-driven (SR, SINDy) and hybrid
variants are able to recover the internal dynamics $\tau_{dyn}$
essentially perfectly.

Comparing training and test performance, we observe that the SR, SINDy
and all hybrid models generalize very well: their test RMSEs closely
match the training RMSEs for every joint.
This is consistent with the fact that the simulator defines a smooth,
low-noise mapping from the feature vector $x$ to $\tau_m$, which can
be represented accurately either as a sparse polynomial or as a compact
symbolic expression.
The hybrid models do not exhibit a clear systematic advantage over
their non-hybrid counterparts in this setting, which is expected since
$\tau_{rbd}$ already matches the true rigid-body component of the
dynamics and the remaining error term $\epsilon$ is relatively small
and structured.

In contrast, the neural-network baselines perform significantly worse.
On the training set, the NN and r-NN models exhibit relative RMSEs
between roughly $0.02$ and $0.17$ depending on the joint, already one
to two orders of magnitude larger than those of the interpretable
models.
On the test set, the degradation is much more pronounced: test RMSEs
for the NN and r-NN models reach values between $0.24$ and $1.36$
across the seven joints, indicating poor generalization despite having
access to the same feature vector $x$.
The hybrid r-NN model slightly improves training errors over the pure
NN but does not consistently reduce test errors, suggesting that, in
this idealized setting, simply adding a neural residual on top of
$\tau_{rbd}$ does not overcome the optimization and generalization
challenges of the black-box model.

\begin{table*}[t]
\centering
\tiny
\setlength{\tabcolsep}{3pt}
\caption{Closed-form models identified by the interpretable methods on a known dynamics model.
We used
$[6.75,\;6.00,\;5.25,\;4.50,\;3.75,\;3.00,\;2.25]^\top$ for the nominal viscous damping coefficients.}
\label{tab:franka-equations}
\begin{tabular}{@{}llp{0.15\textwidth}p{0.35\textwidth}p{0.4\textwidth}@{}}
\hline
 & $j$ & Expected model & Identified model & Coefficients \\
\hline
\multirow{7}{*}{\rotatebox[origin=c]{90}{SR}}
  & 1 &
  \multirow{7}{*}{\eqref{eq:franka-model}} &
  $\xi^{(1)}_1 \tau_{i,1} + \xi^{(1)}_2 \tau_{c,1} + \xi^{(1)}_3 \dot{q}_1$ &
  $\xi^{(1)}_1 = 1,\ \xi^{(1)}_2 = 1,\ \xi^{(1)}_3 = 6.721$ \\
 & 2 & &
  $\xi^{(2)}_1 \tau_{i,2} + \xi^{(2)}_2 \tau_{c,2} + \xi^{(2)}_3 \tau_{g,2} + \xi^{(2)}_4 \dot{q}_2$ &
  $\xi^{(2)}_1 = 1,\ \xi^{(2)}_2 = 1,\ \xi^{(2)}_3 = 1,\ \xi^{(2)}_4 = 5.962$ \\
 & 3 & &
  $\xi^{(3)}_1 \tau_{i,3} + \xi^{(3)}_2 \tau_{c,3} + \xi^{(3)}_3 \tau_{g,3} + \xi^{(3)}_4 \dot{q}_3$ &
  $\xi^{(3)}_1 = 1,\ \xi^{(3)}_2 = 1,\ \xi^{(3)}_3 = 1,\ \xi^{(3)}_4 = 5.235$ \\
 & 4 & &
  $\xi^{(4)}_1 \tau_{i,4} + \xi^{(4)}_2 \tau_{c,4} + \xi^{(4)}_3 \tau_{g,4} + \xi^{(4)}_4 \dot{q}_4$ &
  $\xi^{(4)}_1 = 1,\ \xi^{(4)}_2 = 1,\ \xi^{(4)}_3 = 1,\ \xi^{(4)}_4 = 4.486$ \\
 & 5 & &
  $\xi^{(5)}_1 \tau_{i,5} + \xi^{(5)}_2 \tau_{c,5} + \xi^{(5)}_3 \tau_{g,5} + \xi^{(5)}_4 \dot{q}_5 + \xi^{(5)}_5 \dddot{q}_5$ &
  $\xi^{(5)}_1 \approx 1.001,\ \xi^{(5)}_2 = 1,\ \xi^{(5)}_3 = 1,$\\
 &   & &
  & $\xi^{(5)}_4 \approx 3.741,\ \xi^{(5)}_5 \approx 5.2\times 10^{-5}$ \\
 & 6 & &
  $\xi^{(6)}_1 \tau_{i,6} + \xi^{(6)}_2 \tau_{c,6} + \xi^{(6)}_3 \tau_{g,6} + \xi^{(6)}_4 \dot{q}_6 + \xi^{(6)}_5 \dddot{q}_6$ &
  $\xi^{(6)}_1 \approx 1.001,\ \xi^{(6)}_2 = 1,\ \xi^{(6)}_3 = 1,$\\
 &   & &
  & $\xi^{(6)}_4 \approx 2.993,\ \xi^{(6)}_5 \approx 1.0\times 10^{-4}$ \\
 & 7 & &
  $\xi^{(7)}_1 \tau_{i,7} + \xi^{(7)}_2 \tau_{c,7} + \xi^{(7)}_3 \dot{q}_7$ &
  $\xi^{(7)}_1 = 1,\ \xi^{(7)}_2 = 1,\ \xi^{(7)}_3 = 2.244$ \\
\hline
\multirow{7}{*}{\rotatebox[origin=c]{90}{SINDy}}
  & 1 &
  \multirow{7}{*}{\eqref{eq:franka-model}} &
  $\xi^{(1)}_1 \tau_{i,1} + \xi^{(1)}_2 \tau_{c,1} + \xi^{(1)}_3 \dot{q}_1$ &
  $\xi^{(1)}_1 = 1.001,\ \xi^{(1)}_2 = 1.001,\ \xi^{(1)}_3 = 6.721$ \\
 & 2 & &
  $\xi^{(2)}_1 \tau_{i,2} + \xi^{(2)}_2 \tau_{c,2} + \xi^{(2)}_3 \tau_{g,2} + \xi^{(2)}_4 \dot{q}_2$ &
  $\xi^{(2)}_1 = 1,\ \xi^{(2)}_2 = 1.001,\ \xi^{(2)}_3 = 1,\ \xi^{(2)}_4 = 5.962$ \\
 & 3 & &
  $\xi^{(3)}_1 \tau_{i,3} + \xi^{(3)}_2 \tau_{c,3} + \xi^{(3)}_3 \tau_{g,3} + \xi^{(3)}_4 \dot{q}_3$ &
  $\xi^{(3)}_1 = 1.001,\ \xi^{(3)}_2 = 1.002,\ \xi^{(3)}_3 = 1,\ \xi^{(3)}_4 = 5.234$ \\
 & 4 & &
  $\xi^{(4)}_1 \tau_{i,4} + \xi^{(4)}_2 \tau_{c,4} + \xi^{(4)}_3 \tau_{g,4} + \xi^{(4)}_4 \dot{q}_4$ &
  $\xi^{(4)}_1 = 1,\ \xi^{(4)}_2 = 1.001,\ \xi^{(4)}_3 = 1,\ \xi^{(4)}_4 = 4.486$ \\
 & 5 & &
  $\xi^{(5)}_1 \tau_{i,5} + \xi^{(5)}_2 \tau_{c,5} + \xi^{(5)}_3 \tau_{g,5} + \xi^{(5)}_4 \dot{q}_5$ &
  $\xi^{(5)}_1 = 1.001,\ \xi^{(5)}_2 = 1,\ \xi^{(5)}_3 \approx 0.9997,\ \xi^{(5)}_4 = 3.741$ \\
 & 6 & &
  $\xi^{(6)}_1 \tau_{i,6} + \xi^{(6)}_2 \tau_{c,6} + \xi^{(6)}_3 \tau_{g,6} + \xi^{(6)}_4 \dot{q}_6$ &
  $\xi^{(6)}_1 = 1.001,\ \xi^{(6)}_2 = 1.001,\ \xi^{(6)}_3 = 1,\ \xi^{(6)}_4 = 2.991$ \\
 & 7 & &
  $\xi^{(7)}_1 \tau_{i,7} + \xi^{(7)}_2 \tau_{c,7} + \xi^{(7)}_3 \dot{q}_7$ &
  $\xi^{(7)}_1 = 1.001,\ \xi^{(7)}_2 = 1.001,\ \xi^{(7)}_3 = 2.244$ \\
\hline
\multirow{7}{*}{\rotatebox[origin=c]{90}{r-SR}}
  & 1 &
  \multirow{7}{*}{$\epsilon_j = d_j \dot{q}_j$} &
  $\xi^{(1)}_1 \dot{q}_1$ &
  $\xi^{(1)}_1 = 6.721$ \\
 & 2 & &
  $\xi^{(2)}_1 \dot{q}_2$ &
  $\xi^{(2)}_1 = 5.962$ \\
 & 3 & &
  $\xi^{(3)}_1 \dot{q}_3$ &
  $\xi^{(3)}_1 = 5.235$ \\
 & 4 & &
  $\xi^{(4)}_1 \dot{q}_4$ &
  $\xi^{(4)}_1 = 4.486$ \\
 & 5 & &
  $\xi^{(5)}_1 \dot{q}_5 + \xi^{(5)}_2 \dddot{q}_5$ &
  $\xi^{(5)}_1 = 3.742,\ \xi^{(5)}_2 \approx 5.2\times 10^{-5}$ \\
 & 6 & &
  $\xi^{(6)}_1 \dot{q}_6 + \xi^{(6)}_2 \dddot{q}_6$ &
  $\xi^{(6)}_1 = 2.994,\ \xi^{(6)}_2 \approx 1.0\times 10^{-4}$ \\
 & 7 & &
  $\xi^{(7)}_1 \dot{q}_7$ &
  $\xi^{(7)}_1 = 2.244$ \\
\hline
\multirow{7}{*}{\rotatebox[origin=c]{90}{r-SINDy}}
  & 1 &
  \multirow{7}{*}{$\epsilon_j = d_j \dot{q}_j$} &
  $\xi^{(1)}_1 \dot{q}_1$ &
  $\xi^{(1)}_1 = 6.721$ \\
 & 2 & &
  $\xi^{(2)}_1 \dot{q}_2$ &
  $\xi^{(2)}_1 = 5.962$ \\
 & 3 & &
  $\xi^{(3)}_1 \dot{q}_3$ &
  $\xi^{(3)}_1 = 5.235$ \\
 & 4 & &
  $\xi^{(4)}_1 \dot{q}_4$ &
  $\xi^{(4)}_1 = 4.486$ \\
 & 5 & &
  $\xi^{(5)}_1 \dot{q}_5$ &
  $\xi^{(5)}_1 = 3.741$ \\
 & 6 & &
  $\xi^{(6)}_1 \dot{q}_6$ &
  $\xi^{(6)}_1 = 2.991$ \\
 & 7 & &
  $\xi^{(7)}_1 \dot{q}_7$ &
  $\xi^{(7)}_1 = 2.244$ \\
\hline
\multirow{7}{*}{\rotatebox[origin=c]{90}{r-SINDy-SR}}
  & 1 &
  \multirow{7}{*}{$\epsilon^{(2)}_j = 0$} &
  $\xi^{(1)}_1 \ddot{q}_1$ &
  $\xi^{(1)}_1 \approx 2.92\times 10^{-3}$ \\
 & 2 & &
  $\xi^{(2)}_1 \ddot{q}_2$ &
  $\xi^{(2)}_1 \approx 2.38\times 10^{-3}$ \\
 & 3 & &
  $\xi^{(3)}_1 \ddot{q}_3 + \xi^{(3)}_2 \dot{q}_1$ &
  $\xi^{(3)}_1 \approx 2.74\times 10^{-3},\ \xi^{(3)}_2 \approx -2.74\times 10^{-3}$ \\
 & 4 & &
  $\xi^{(4)}_1 \dddot{q}_4$ &
  $\xi^{(4)}_1 \approx 9.62\times 10^{-5}$ \\
 & 5 & &
  $\xi^{(5)}_1 \ddot{q}_5$ &
  $\xi^{(5)}_1 \approx 2.04\times 10^{-3}$ \\
 & 6 & &
  $\xi^{(6)}_1 \ddot{q}_6$ &
  $\xi^{(6)}_1 \approx 1.53\times 10^{-3}$ \\
 & 7 & &
  $\xi^{(7)}_1 \ddot{q}_7$ &
  $\xi^{(7)}_1 \approx 1.03\times 10^{-3}$ \\
\hline
\end{tabular}
\vspace{-0.5cm}
\end{table*}

Table~\ref{tab:franka-equations} reports the closed-form expressions identified by the interpretable methods and enables a direct comparison with the target structure in~\eqref{eq:franka-model} and the error decomposition in~\eqref{eq:dyn-err}. 
For the purely data-driven SR and SINDy models, the learned expressions closely match the desired superposition of inertial, Coriolis/centrifugal, gravity, and viscous contributions.
Across all joints, the coefficients multiplying $\tau_{i,j}$, $\tau_{c,j}$, and $\tau_{g,j}$ are essentially equal to one (up to small $\mathcal{O}(10^{-3})$ deviations), indicating that both methods correctly recover the known rigid-body components from data.
Moreover, the estimated viscous coefficients $\xi^{(j)}$ on $\dot{q}_j$ agree very well with the nominal damping coefficients $d_j$, with relative discrepancies below $1\%$ for all joints.
For joints~5 and~6, SR additionally introduces jerk terms $\dddot{q}_j$ with very small magnitude (on the order of $10^{-5}$–$10^{-4}$), consistent with the negligible higher-order corrections suggested by the low RMSE values in Tables~\ref{tab:franka-rmse_train} and~\ref{tab:franka-rmse_test}.

The hybrid models further confirm that the dominant unmodeled dynamics in this setup are well captured by a simple viscous term.
Both r-SR and r-SINDy recover $\epsilon_j$ as a scalar multiple of $\dot{q}_j$ that is numerically indistinguishable from $d_j$, with only very small jerk corrections for joints~5 and~6 in the r-SR case.
In contrast, the r-SINDy-SR model, which refines the hybrid RBD--SINDy estimate, learns only very small residual structures in terms of accelerations $\ddot{q}_j$ (and a single tiny cross term involving $\dot{q}_1$ for joint~3), with coefficients on the order of $10^{-3}$ or smaller.
Taken together, these results show that once the rigid-body model and a joint-wise viscous damping term are accounted for, there is little systematic structure remaining in the residual, and the interpretable methods converge to compact, physically meaningful expressions that are consistent with the known simulated dynamics.

\subsubsection{Discussion}

Overall, these results show that in a noise-free environment with a
known rigid-body model, interpretable sparsity-promoting methods
(SR, SINDy and their hybrid variants) can recover a highly accurate
model of $\tau_{dyn}$ and the error term $\epsilon$ in
\eqref{eq:dyn-err}, whereas standard neural-network baselines struggle
to match their accuracy and generalization.
The symbolic models not only achieve low training and test RMSE across
all joints, but also recover coefficients that are numerically
consistent with the known viscous damping and rigid-body terms, and
expose any remaining structure in the residual as small, easily
interpretable corrections (e.g., weak jerk or acceleration terms).

By contrast, the pure NN and hybrid RBD--NN baselines exhibit clear
signs of overfitting in this setting, attaining low training error but
substantially worse test performance and offering no direct insight
into the underlying physical structure of the dynamics.
Taken together, the simulation study validates that, when a
reasonable mechanistic model is available and the data are sufficiently
informative, interpretable sparse regression can serve as a powerful
tool to identify both $\tau_{dyn}$ and $\epsilon$ in
\eqref{eq:dyn-err}, combining prediction accuracy with physically
meaningful, human-readable models that can be inspected, validated,
and potentially modified by a designer.

\subsection{Analysis on real world data from WAM robot arm}


The previous section provided evidence that interpretable methods can be used to discover dynamics models from data. 
Also, the experimental pipeline developed was validated.
In this section, we wish to apply and analyze these interpretable methods on real world data collected from the 7-DoF WAM robot arm. 

\subsubsection{Dataset}

The dataset for the experiments in this section was sourced from the work of Sousa and Cortes\~{a}o~\cite{sousa2014}, available at
\href{https://github.com/cdsousa/wam7_dyn_ident}{github.com/cdsousa/wam7\_dyn\_ident}.

Estimating time derivatives from noisy joint trajectories is delicate, since standard finite-difference schemes tend to amplify measurement noise and can severely degrade the quality of learned models. 
For the WAM data, we therefore computed joint velocity, acceleration, and jerk using a smoothed finite-difference method based on the total-variation-regularized differentiation framework of Chartrand~\cite{chartrand2011numerical}. In this approach, the derivative signal is obtained as the minimizer of a small variational problem that balances a least-squares fit to the noisy measurements with a total-variation penalty on the derivative, effectively enforcing a piecewise-smooth derivative that suppresses high-frequency noise while still allowing sharp changes. We apply this procedure independently to each joint trajectory before constructing the feature matrices.

\begin{figure*}[ht]
    \centering
    \includegraphics[width=\linewidth]{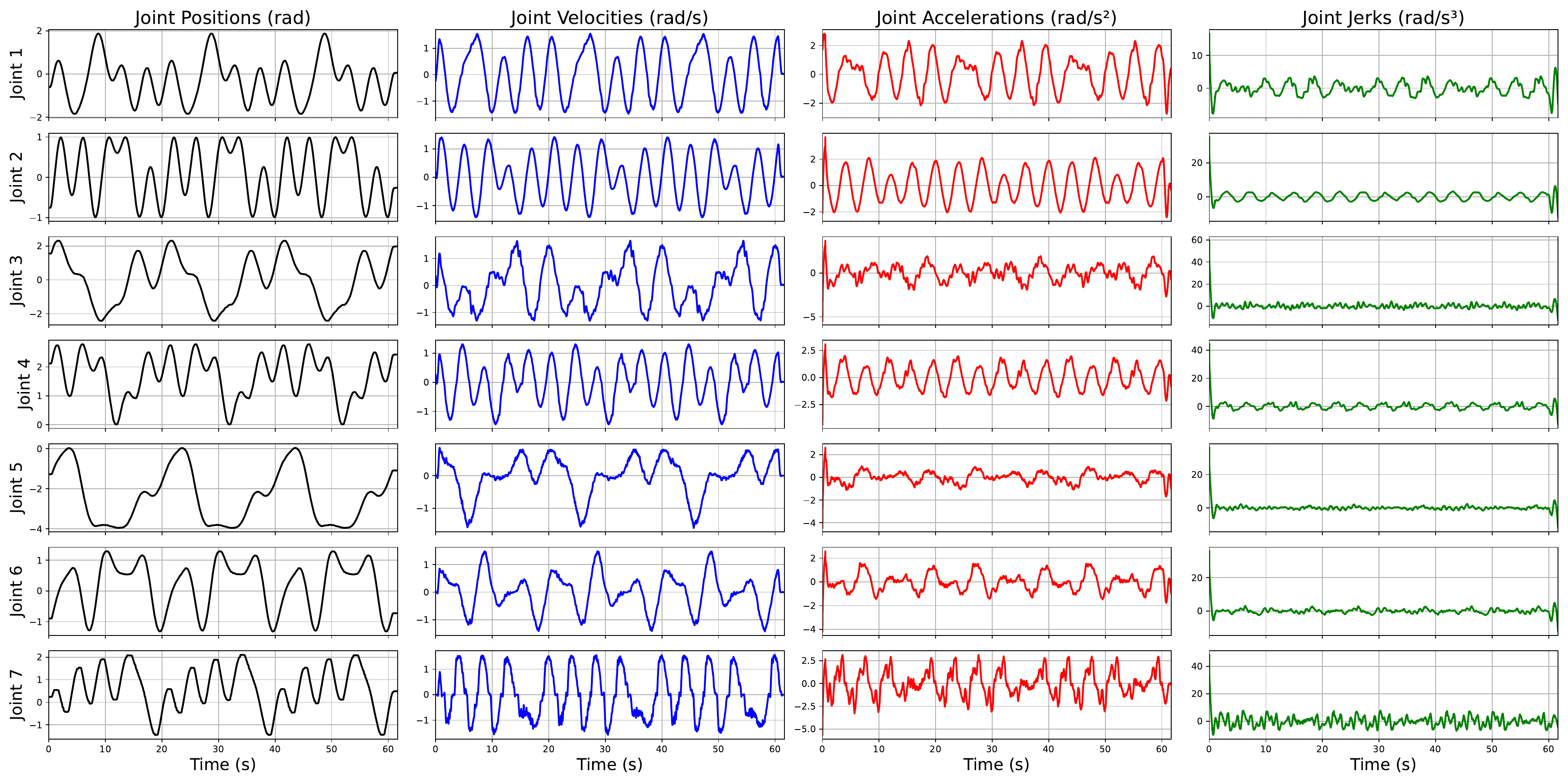}
    \caption{An example of one of the trajectories provided by Sousa and Cortes\~{a}o~\cite{sousa2014}, along with estimated velocities, accelerations, and jerk. Estimates of the various time derivatives were found using the method proposed by Chartrand~\cite{chartrand2011numerical}.}
    \label{fig:wam-traj}
    \vspace{-0.5cm}
\end{figure*}

The dataset provided by Sousa and Cortes\~{a}o~\cite{sousa2014} contains four robot trajectories, all containing about one minute of robot motion collected at a sampling frequency of 1 kHz. 
We used three trajectories as part of the training dataset and one as test dataset.
An example of the joint positions provided in the dataset, along with the estimated velocities, accelerations and jerk (estimated using the smoothed finite differencing method described above) is shown in Figure~\ref{fig:wam-traj}.

\subsubsection{Results}

\begin{table}[t]
\centering
\small 
\setlength{\tabcolsep}{3pt} 
\caption{WAM training set RMSE per joint.}
\label{tab:wam-rmse_train}
\begin{tabular}{@{}cccccccc@{}}
\hline
Joint & SR & SINDy & r-SR & r-SINDy & r-SINDy-SR & NN & r-NN \\
\hline
1 & 0.410 & \textbf{0.103} & 0.477 & \textbf{0.103} & \textbf{0.103} & 0.121 & 0.132 \\
2 & 0.143 & \textbf{0.027} & 0.182 & \textbf{0.027} & \textbf{0.027} & 0.030 & 0.030 \\
3 & 0.437 & 0.057 & 0.570 & \textbf{0.057} & \textbf{0.057} & 0.081 & 0.081 \\
4 & 0.423 & \textbf{0.039} & 0.422 & \textbf{0.039} & \textbf{0.039} & 0.056 & 0.058 \\
5 & 0.621 & \textbf{0.177} & 0.645 & 0.177 & 0.177 & 0.329 & 0.330 \\
6 & 0.487 & \textbf{0.076} & 0.556 & \textbf{0.076} & \textbf{0.076} & 0.213 & 0.213 \\
7 & 0.403 & \textbf{0.090} & 0.558 & 0.090 & 0.090 & 0.322 & 0.324 \\
\hline
\end{tabular}

\caption{WAM test set RMSE per joint.}
\label{tab:wam-rmse_test}
\begin{tabular}{@{}cccccccc@{}}
\hline
Joint & SR & SINDy & r-SR & r-SINDy & r-SINDy-SR & NN & r-NN \\
\hline
1 & \textbf{0.410} & 18.566 & 0.441 & 18.566 & 18.566 & 0.733 & 0.681 \\
2 & \textbf{0.183} & 4.106 & 0.230 & 4.106 & 4.106 & 0.470 & 0.472 \\
3 & \textbf{0.483} & 7.782 & 0.540 & 7.463 & 7.463 & 0.876 & 0.773 \\
4 & \textbf{0.453} & 9.423 & 0.458 & 9.423 & 9.423 & 0.805 & 0.854 \\
5 & 0.476 & 60.189 & \textbf{0.475} & 59.389 & 59.388 & 0.942 & 1.243 \\
6 & 0.548 & 13.280 & \textbf{0.547} & 13.160 & 13.160 & 0.822 & 0.967 \\
7 & \textbf{0.437} & 27.754 & 0.580 & 27.905 & 27.905 & 1.246 & 1.282 \\
\hline
\end{tabular}
\vspace{-0.5cm}
\end{table}

Table~\ref{tab:wam-rmse_train} reports the per-joint relative RMSE
on the WAM training set, and Table~\ref{tab:wam-rmse_test} shows the
corresponding errors on the test set.
On the training data, the SINDy-based models (SINDy, r-SINDy, r-SINDy-SR)
achieve the lowest RMSE for every joint, with values between roughly
$0.03$ and $0.18$, i.e., substantially smaller than those of SR and r-SR
(about $0.14$–$0.65$) and somewhat better than the NN and r-NN baselines
(about $0.03$–$0.33$).
Thus, in terms of in-sample fit, the sparse-regression models remain
very effective at capturing the observed joint torques on this system.

The generalization behavior on the WAM test set, however, is dramatically
different across model classes.
The SINDy and SINDy-based hybrid models suffer from severe overfitting:
their test RMSEs increase by two to three orders of magnitude compared
to training, reaching values between approximately $4$ and $60$ depending
on the joint.
In contrast, the SR and r-SR models exhibit much more stable behaviour,
with test RMSEs in the range $0.18$–$0.58$ that remain close to their
training errors.
For most joints, the best-performing method on the WAM test set
(highlighted in bold in Table~\ref{tab:wam-rmse_test}) is therefore SR
or r-SR, indicating that the symbolic-regression models provide the most
robust out-of-sample performance in this setting.

The neural-network baselines occupy an intermediate regime.
On the training set, NN and r-NN attain errors comparable to or slightly
larger than the SINDy family, but substantially lower than SR and r-SR.
On the test set, their RMSEs increase to roughly $0.47$–$1.28$, which is
still an order of magnitude smaller than the SINDy-based models but
consistently worse than SR and r-SR across all joints.
The hybrid r-NN slightly improves over the pure NN for some joints
(e.g., joint~1) but degrades performance on others, and does not close
the gap to the best SR-based models.

The SR approach produced the following equations.
\begin{align}
    \widehat{\tau}^{SR}_{m,1} & =  \ddot{q}_1 + \dot{q}_1(\xi_1^{(1)} - \dot{q}_1^2)\\
    \widehat{\tau}^{SR}_{m,2} & =  \ddot{q}_2 - \ddot{q}_3 + \dot{q}_2(\dot{q}_5 + \xi_2^{(1)}) - \dot{q}_3^2 \\
    & \quad - ((\xi_2^{(2)} - q_4)q_2\xi_2^{(3)} - q_6 + \tau_{g,3}) \nonumber\\
    \widehat{\tau}^{SR}_{m,3} & = \dot{q}_3 + (\ddot{q}_1 + q_3q_4(\xi_3^{(1)}\ddot{q}_1 + \xi_{3}^{(2)}))(q_2 + \xi_3^{(3)})\\
    \widehat{\tau}^{SR}_{m,4} & = (q_4 - \xi_4^{(1)})\xi_4^{(2)}q_4 + \dot{q}_4 + \xi_4^{(3)}\\
    \widehat{\tau}^{SR}_{m,5} & = \xi_5^{(1)}\dot{q}_5\\
    \widehat{\tau}^{SR}_{m,6} & = (\xi_6^{(1)} - \tau_{g,6})(\dot{q}_6 + (\ddot{q}_6 + \dot{q}_3)\xi_6^{(2)})\\
    \widehat{\tau}^{SR}_{m,7} & = \dot{q}_7(\dot{q}_7^2(\xi_7^{(1)}\dot{q}_7^2 - \xi_7^{(2)}) + \xi_7^{(3)})
\end{align}
where 
$\xi_1^{(1)} = 4.723089$,
$\xi_2{(1)} = 2.9080215$,
$\xi_2^{(2)} = 7.1627355$,
$\xi_2^{(3)} = 4.039454$,
$\xi_3^{(1)} = 0.47727528$,
$\xi_3^{(2)} = 1.7755635$,
$\xi_3^{(3)} = 0.11157574$,
$\xi_4^{(1)} = 3.4464025$,
$\xi_4^{(2)} = 1.9739965$,
$\xi_4^{(3)} = 1.8875742$,
$\xi_5^{(1)} = 0.14579542$,
$\xi_6^{(1)} = 0.26862046$,
$\xi_6^{(2)} = -0.12590475$,
$\xi_7^{(1)} = 0.05014042$,
$\xi_7^{(2)} = 0.17362133$,
$\xi_7^{(3)} = 0.19907075$.

The r-SR approach produced the following equations.
\begin{align}
    \widehat{\epsilon}^{\text{r-SR}}_{m,1} & = -\ddot{q}_1 - \tau_{g,1} + \xi_1^{(1)}\dot{q}_1\\
    \widehat{\epsilon}^{\text{r-SR}}_{m,2} & = (\dot{q}_2 + (\xi_2^{(1)} - q_2)(\xi_2^{(2)}q_4 - \xi_2^{(3)}) - \xi_2^{(4)})\xi_2^{(5)} \\
    \widehat{\epsilon}^{\text{r-SR}}_{m,3} & = q_2q_3\xi_3^{(1)} - \tau_{g,3} + \dot{q}_3\\
    \widehat{\epsilon}^{\text{r-SR}}_{m,4} & = \dot{q}_4 + (q_4 - \xi_4^{(1)})(\xi_4^{(2)}q_4 - \xi_4^{(3)}) \\
    \widehat{\epsilon}^{\text{r-SR}}_{m,5} & = -\tau_{g,5} + \xi_5^{(1)}\dot{q}_5\\
    \widehat{\epsilon}^{\text{r-SR}}_{m,6} & = \xi_6^{(1)}\dot{q}_6\\
    \widehat{\epsilon}^{\text{r-SR}}_{m,7} & = \xi_7^{(1)}\dot{q}_7
\end{align}
where 
$\xi_1^{(1)} = 3.2723246$,
$\xi_2^{(1)} = 0.016012268$,
$\xi_2^{(2)} = -1.023108$,
$\xi_2^{(3)} = -8.626232$,
$\xi_2^{(4)} = 0.50142884$,
$\xi_2^{(5)} = 3.391348$,
$\xi_3^{(1)} = 2.58501$,
$\xi_4^{(1)} = 3.1465542$,
$\xi_4^{(2)} = 1.9603553$,
$\xi_4^{(3)} = 0.58640337$,
$\xi_5^{(1)} = 0.14822637$,
$\xi_6^{(1)} = 0.24277882$,
$\xi_7^{(1)} = 0.07425557$.

The methods using SINDy produced models that are too long to write in the page.

Beyond the quantitative comparison in Tables~\ref{tab:wam-rmse_train} and~\ref{tab:wam-rmse_test}, the closed-form SR and r-SR expressions for the WAM arm provide additional qualitative insight into the structure of the learned dynamics. For several joints, the models recover familiar physical motifs: joints~1, 5, 6, and 7 are dominated by velocity-dependent terms that resemble viscous or nonlinear friction laws (linear in $\dot{q}_j$ for joints~5–7 in the r-SR residuals, and cubic/quintic in $\dot{q}_1$ and $\dot{q}_7$ in the full SR model), while joint~4 features a clear spring-like dependence on $q_4$ of the form $(q_4 - \text{offset})\,q_4$ combined with a viscous term in $\dot{q}_4$ and a constant bias torque. These patterns are consistent with the intuition that the nominal rigid-body model accounts for the bulk of the inertial and gravitational effects, whereas the residuals capture joint-local friction, compliance, and offset torques that arise from the WAM’s cable transmissions and hardware-specific characteristics. The presence of explicit $\tau_{g,j}$ terms in several SR and r-SR equations—for instance, $\tau_{g,3}$ appearing in the models for joints~2 and~3, or $\tau_{g,5}$ and $\tau_{g,6}$ modulating the residuals of joints~5 and~6—further indicates that the symbolic learner makes systematic use of the available gravity regressor, effectively re-weighting or partially ``undoing'' it where the analytical model and the real robot diverge.

At the same time, the WAM expressions also expose more complex structures that go beyond the canonical RBD-plus-viscous-friction picture and would be difficult to anticipate a priori, yet remain compact and interpretable. In particular, the SR model for joints~2 and~3 contains configuration- and coupling-dependent factors such as $(\xi_2^{(2)} - q_4)q_2$, and mixed dependencies on $q_4$, $q_6$, and $\tau_{g,3}$, while the r-SR residual for joint~2 exhibits a product $(\dot{q}_2 + (\xi_2^{(1)} - q_2)(\xi_2^{(2)}q_4 - \xi_2^{(3)}) - \xi_2^{(4)})\xi_2^{(5)}$. These terms suggest nontrivial cross-joint couplings and configuration-dependent biases that are not explicitly encoded in the nominal RBD model, and may be absorbing effects of cable routing, structural compliance, or sensor/actuator nonlinearities. Similarly, the SR expression for joint~6 includes a gravity-modulated friction term $(\xi_6^{(1)} - \tau_{g,6})(\dot{q}_6 + (\ddot{q}_6 + \dot{q}_3)\xi_6^{(2)})$, indicating that the effective damping on joint~6 depends on both posture (through $\tau_{g,6}$) and motion of another joint (via $\dot{q}_3$), which goes beyond standard jointwise viscous friction models. While some of these couplings may reflect identification artifacts or dataset-specific correlations rather than entirely new physical effects, the fact that they arise as low-complexity terms and still generalize better than both SINDy and neural networks on held-out trajectories makes them valuable hypotheses for further mechanical analysis and model refinement.

Finally, we note that all SR and r-SR models on the WAM dataset were learned with a deliberately simple PySR setup: the operator set was restricted to the binary operators $\{+,-,\times\}$, and all remaining options (evolutionary strategy, complexity--accuracy trade-off, population sizes, etc.) were left at their default values, with no task-specific hyperparameter tuning. In other words, the discovered spring-like terms, cross-joint couplings, and gravity-modulated friction laws emerge under a generic, off-the-shelf configuration. It is plausible that a more targeted search---for example, adjusting the complexity penalty, enriching the operator set with piecewise or non-smooth functions, or biasing the search toward cross-joint features---could either simplify some of the more intricate couplings found here or uncover additional, more physically structured components (such as clearer Coulomb-like friction terms or smoother approximations of stick--slip effects) while retaining or further improving generalization on held-out trajectories.

\begin{figure*}[t]
  \centering
  \begin{subfigure}[b]{0.48\textwidth}
    \centering
    \includegraphics[width=\linewidth, clip, trim={0cm 0cm 0cm 1.25cm}]{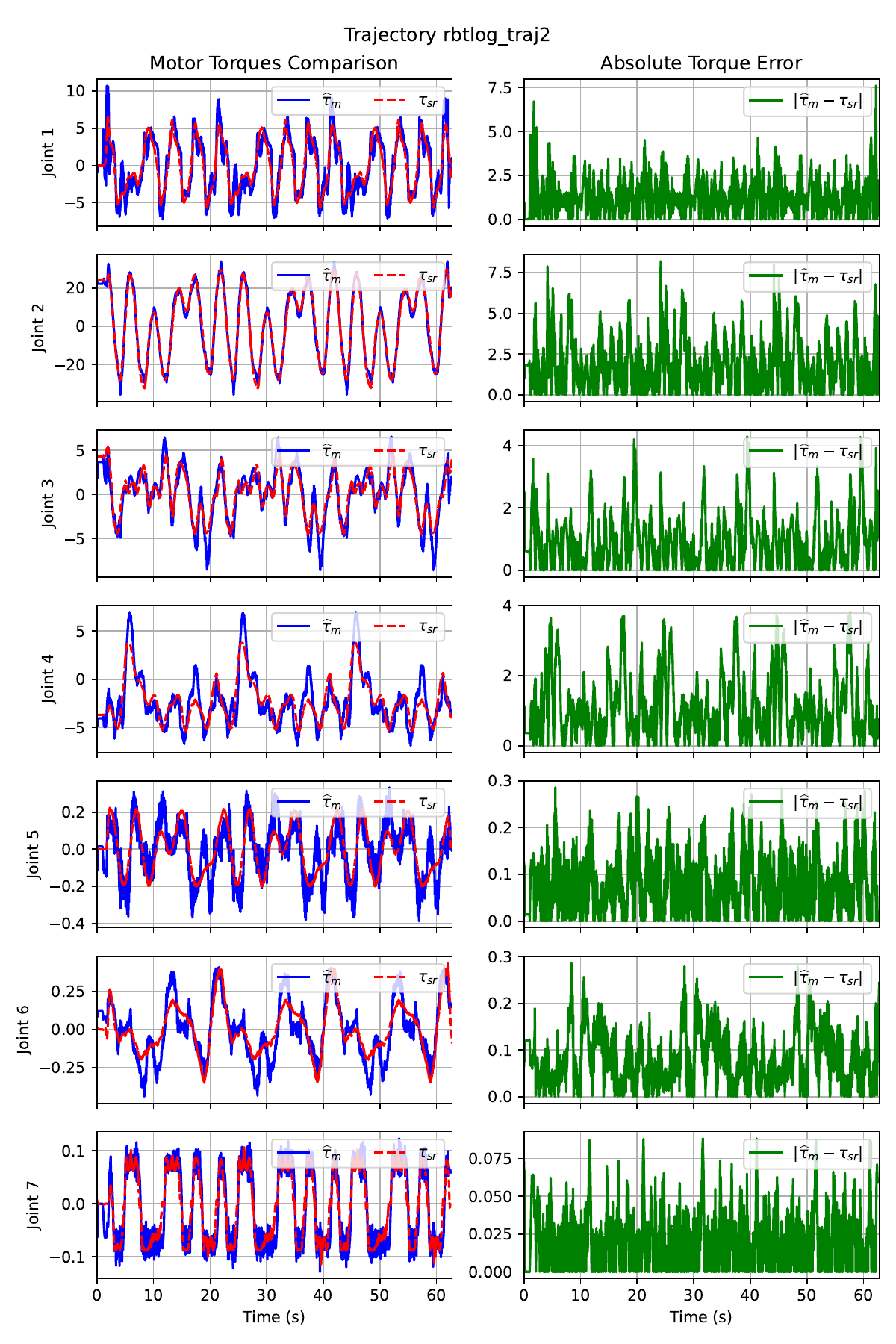}
    \caption{Evaluation on training set.}
    \label{fig:one}
  \end{subfigure}
  \hfill
  \begin{subfigure}[b]{0.48\textwidth}
    \centering
    \includegraphics[width=\linewidth, clip, trim={0cm 0cm 0cm 1.25cm}]{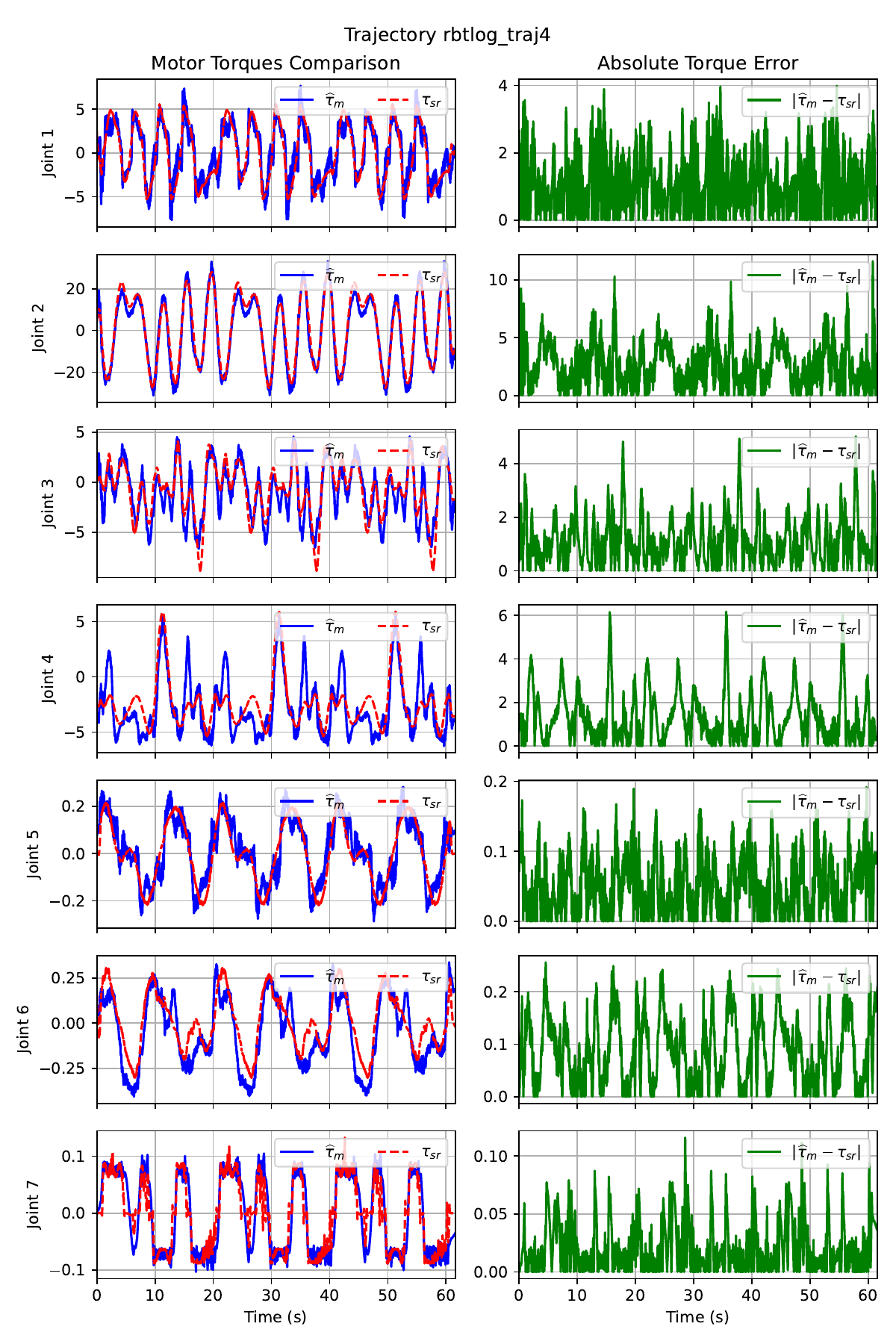}
    \caption{Evaluation on test set.}
    \label{fig:two}
  \end{subfigure}
  \caption{Evaluations of our trained model using SR on the real world data collected from the 7-DoF WAM robot arm~\cite{sousa2014}.}
  \label{fig:combined}
  \vspace{-0.5cm}
\end{figure*}

\subsubsection{Discussion}

Overall, the WAM results highlight a clear gap between in-sample fit and
out-of-sample reliability.
SINDy and its hybrid variants achieve the lowest training errors but fail
to generalize, with test RMSEs increasing by several orders of magnitude
and yielding very long expressions that appear to track trajectory-specific
artefacts rather than underlying mechanics.
In contrast, SR and r-SR trade some training accuracy for substantially
better generalization, attaining the lowest test errors across most joints.

As discussed above, the closed-form SR and r-SR models remain compact and
physically interpretable, capturing friction-like, spring-like, and
gravity-modulated effects as well as nontrivial cross-joint couplings.
Importantly, all of these SR and r-SR models were obtained with a simple,
off-the-shelf PySR configuration restricted to the operators
$\{+,-,\times\}$ and default evolutionary settings, with no
task-specific hyperparameter tuning.
This suggests that even a modest symbolic-regression setup can already
uncover meaningful structure in realistic robot data.
The limited benefit of hybridizing neural networks with $\tau_{rbd}$ and
the poor robustness of the SINDy variants together underline that, in
this more realistic setting where the analytical model and function
libraries are only approximate, the choice of model class and
regularization is at least as important as raw approximation power.

\section{Conclusion}

We proposed interpretable, data-driven residual models of robot dynamics that augment an analytical rigid-body dynamics (RBD) model with a learned error term $\epsilon$. Using symbolic regression (SR) and SINDy-style sparse regression on joint-space features, we obtained compact expressions that, in simulation on a Franka arm with known dynamics, recover inertial, Coriolis, gravity, and viscous terms with relative errors of order $10^{-3}$. In this setting, SR/SINDy and their RBD hybrids match the true dynamics with strong train--test generalization and outperform neural-network baselines while remaining fully interpretable. On real data from a 7-DoF WAM arm, SR models generalize best and consistently outperform both SINDy and neural networks on held-out data, revealing physically meaningful structure such as viscous and nonlinear friction, spring-like terms, gravity reweightings, and low-dimensional cross-joint couplings.

\subsection{Limitations}

Our method assumes an accurate RBD model and sufficiently clean joint-level measurements. Models are trained offline on pre-collected data and evaluated via one-step torque prediction, so their closed-loop impact on performance and safety is unknown. We restrict the feature libraries and function classes to hand-designed polynomials, a basic operator set $\{+,-,\times\}$, and per-joint models, which may miss richer friction effects and higher-order couplings. Derivative-based features (velocities, accelerations, jerk) rely on numerical differentiation; although we use smoothed, total-variation-regularized finite differences~\cite{chartrand2011numerical}, derivative quality and some learned terms may still be affected by noise, sampling rates, and sensor biases.

\subsection{Future Work}

Future work will embed interpretable residual models into control and learning pipelines, e.g., as components in impedance or model-predictive controllers, as structured priors in model-based or hybrid reinforcement learning, and as virtual force/torque sensors for contact-rich tasks. We also aim to relax reliance on a fixed RBD model by jointly identifying rigid-body parameters and residuals, and by enriching SR/SINDy libraries with more physics-informed, possibly non-smooth operators while preserving sparsity. Finally, we plan to study transfer across robots and data regimes, leveraging shared interpretable structure to generalize dynamics models and support scalable, force-aware extensions of vision--language--action frameworks.

\bibliographystyle{unsrtnat}
\bibliography{references}

@misc{Scholl25,
      title={Interpretable Robotic Friction Learning via Symbolic Regression}, 
      author={Philipp Scholl and Alexander Dietrich and Sebastian Wolf and Jinoh Lee and Alin-Albu Schäffer and Gitta Kutyniok and Maged Iskandar},
      year={2025},
      eprint={2505.13186},
      archivePrefix={arXiv},
      primaryClass={cs.RO},
      url={https://arxiv.org/abs/2505.13186}, 
}

@ARTICLE{Swevers97,
  author={Swevers, J. and Ganseman, C. and Tukel, D.B. and de Schutter, J. and Van Brussel, H.},
  journal={IEEE Transactions on Robotics and Automation}, 
  title={Optimal robot excitation and identification}, 
  year={1997},
  volume={13},
  number={5},
  pages={730-740},
  doi={10.1109/70.631234}}

@inproceedings{tian2024excitation,
  title={Excitation trajectory optimization for dynamic parameter identification using virtual constraints in hands-on robotic system},
  author={Tian, Huanyu and Huber, Martin and Mower, Christopher E and Han, Zhe and Li, Changsheng and Duan, Xingguang and Bergeles, Christos},
  booktitle={2024 IEEE International Conference on Robotics and Automation (ICRA)},
  pages={11605--11611},
  year={2024},
  organization={IEEE}
}

@article{chartrand2011numerical,
  title={Numerical differentiation of noisy, nonsmooth data},
  author={Chartrand, Rick},
  journal={International Scholarly Research Notices},
  volume={2011},
  number={1},
  pages={164564},
  year={2011},
  publisher={Wiley Online Library}
}

@misc{pybullet,
  title={Pybullet, a python module for physics simulation for games, robotics and machine learning},
  author={Coumans, Erwin and Bai, Yunfei},
  year={2016}
}

@INPROCEEDINGS{pinocchio,
  author={Carpentier, Justin and Saurel, Guilhem and Buondonno, Gabriele and Mirabel, Joseph and Lamiraux, Florent and Stasse, Olivier and Mansard, Nicolas},
  booktitle={2019 IEEE/SICE International Symposium on System Integration (SII)}, 
  title={The Pinocchio C++ library : A fast and flexible implementation of rigid body dynamics algorithms and their analytical derivatives}, 
  year={2019},
  volume={},
  number={},
  pages={614-619},
  doi={10.1109/SII.2019.8700380}}

@misc{MoralesAlvarado24,
      title={Symbolic regression for precision LHC physics}, 
      author={Manuel Morales-Alvarado and Daniel Conde and Josh Bendavid and Veronica Sanz and Maria Ubiali},
      year={2024},
      eprint={2412.07839},
      archivePrefix={arXiv},
      primaryClass={hep-ph},
      url={https://arxiv.org/abs/2412.07839}, 
}

@INPROCEEDINGS{Khorshidi25,
  author={Khorshidi, Shahram and Dawood, Murad and Nederkorn, Benno and Bennewitz, Maren and Khadiv, Majid},
  booktitle={2025 IEEE International Conference on Robotics and Automation (ICRA)}, 
  title={Physically-Consistent Parameter Identification of Robots in Contact}, 
  year={2025},
  volume={},
  number={},
  pages={677-683},
  doi={10.1109/ICRA55743.2025.11128710}}

@inproceedings{rt2,
  title     = {RT-2: Vision-Language-Action Models Transfer Web Knowledge to Robotic Control},
  author    = {Zitkovich, Brianna and Yu, Tianhe and Xu, Sichun and Xu, Peng and Xiao, Ted and Xia, Fei and Wu, Jialin and Wohlhart, Paul and Welker, Stefan and Wahid, Ayzaan and Vuong, Quan and Vanhoucke, Vincent and Tran, Huong and Soricut, Radu and Singh, Anikait and Singh, Jaspiar and Sermanet, Pierre and Sanketi, Pannag R. and Salazar, Grecia and Ryoo, Michael S. and Reymann, Krista and Rao, Kanishka and Pertsch, Karl and Mordatch, Igor and Michalewski, Henryk and Lu, Yao and Levine, Sergey and Lee, Lisa and Lee, Tsang-Wei Edward and Leal, Isabel and Kuang, Yuheng and Kalashnikov, Dmitry and Julian, Ryan and Joshi, Nikhil J. and Irpan, Alex and Ichter, Brian and Hsu, Jasmine and Herzog, Alexander and Hausman, Karol and Gopalakrishnan, Keerthana and Fu, Chuyuan and Florence, Pete and Finn, Chelsea and Dubey, Kumar Avinava and Driess, Danny and Ding, Tianli and Choromanski, Krzysztof Marcin and Chen, Xi and Chebotar, Yevgen and Carbajal, Justice and Brown, Noah and Brohan, Anthony and Gonzalez Arenas, Montserrat and Han, Kehang},
  booktitle = {Proceedings of The 7th Conference on Robot Learning},
  series    = {Proceedings of Machine Learning Research},
  volume    = {229},
  pages     = {2165--2183},
  year      = {2023},
  publisher = {PMLR},
  url       = {https://proceedings.mlr.press/v229/zitkovich23a.html}
}

@article{openxembodiment,
  title         = {Open X-Embodiment: Robotic Learning Datasets and {RT-X} Models},
  author        = {Open X.-Embodiment Collaboration and Padalkar, Abhishek and Pooley, Acorn and Jain, Ajinkya and Bewley, Alex and Herzog, Alexander and Irpan, Alex and Khazatsky, Alexander and Raj, Anant and Singh, Anikait and Brohan, Anthony and Raffin, Antonin and Wahid, Ayzaan and Burgess-Limerick, Ben and Kim, Beomjoon and Sch{\"o}lkopf, Bernhard and Ichter, Brian and Lu, Cewu and Xu, Charles and Finn, Chelsea and Xu, Chenfeng and Chi, Cheng and Huang, Chenguang and Chan, Christine and Pan, Chuer and Fu, Chuyuan and Devin, Coline and Driess, Danny and Pathak, Deepak and Shah, Dhruv and B{\"u}chler, Dieter and Kalashnikov, Dmitry and Sadigh, Dorsa and Johns, Edward and Ceola, Federico and Xia, Fei and Stulp, Freek and Zhou, Gaoyue and Sukhatme, Gaurav S. and Salhotra, Gautam and Yan, Ge and Schiavi, Giulio and Kahn, Gregory and Su, Hao and Fang, Haoshu and Shi, Haochen and Amor, Heni Ben and Christensen, Henrik I. and Furuta, Hiroki and Walke, Homer and Fang, Hongjie and Mordatch, Igor and Radosavovic, Ilija and others},
  journal       = {CoRR},
  volume        = {abs/2310.08864},
  year          = {2023},
  doi           = {10.48550/arXiv.2310.08864},
  url           = {https://arxiv.org/abs/2310.08864},
  eprint        = {2310.08864},
  archivePrefix = {arXiv}
}

@article{openvla,
  title         = {OpenVLA: An Open-Source Vision-Language-Action Model},
  author        = {Kim, Moo Jin and Pertsch, Karl and Karamcheti, Siddharth and Xiao, Ted and Balakrishna, Ashwin and Nair, Suraj and Rafailov, Rafael and Foster, Ethan P. and Lam, Grace and Sanketi, Pannag R. and Vuong, Quan and Kollar, Thomas and Burchfiel, Benjamin and Tedrake, Russ and Sadigh, Dorsa and Levine, Sergey and Liang, Percy and Finn, Chelsea},
  journal       = {CoRR},
  volume        = {abs/2406.09246},
  year          = {2024},
  doi           = {10.48550/arXiv.2406.09246},
  url           = {https://arxiv.org/abs/2406.09246},
  eprint        = {2406.09246},
  archivePrefix = {arXiv}
}

@article{xie25forceful,
  title         = {Towards Forceful Robotic Foundation Models: a Literature Survey},
  author        = {Xie, William and Correll, Nikolaus},
  journal       = {CoRR},
  volume        = {abs/2504.11827},
  year          = {2025},
  doi           = {10.48550/arXiv.2504.11827},
  url           = {https://arxiv.org/abs/2504.11827},
  eprint        = {2504.11827},
  archivePrefix = {arXiv}
}

@inproceedings{Fankhauser13,
  author    = {P. Fankhauser and M. Hutter and C. Gehring and M. Bloesch and
               M. A. Hoepflinger and R. Siegwart},
  title     = {Reinforcement Learning of Single Legged Locomotion},
  booktitle = {2013 IEEE/RSJ International Conference on Intelligent Robots and Systems (IROS)},
  pages     = {188--193},
  year      = {2013},
  publisher = {IEEE},
  doi       = {10.1109/IROS.2013.6696363}
}

@article{Lee20,
  author  = {Joonho Lee and Jemin Hwangbo and Lorenz Wellhausen and Vladlen Koltun and Marco Hutter},
  title   = {Learning Quadrupedal Locomotion over Challenging Terrain},
  journal = {Science Robotics},
  volume  = {5},
  number  = {47},
  pages   = {eabc5986},
  year    = {2020},
  doi     = {10.1126/scirobotics.abc5986}
}

@inproceedings{Rudin22,
  author    = {Nikita Rudin and David Hoeller and Philipp Reist and Marco Hutter},
  title     = {Learning to Walk in Minutes Using Massively Parallel Deep Reinforcement Learning},
  booktitle = {Proceedings of the 5th Conference on Robot Learning},
  series    = {Proceedings of Machine Learning Research},
  volume    = {164},
  pages     = {91--100},
  year      = {2022},
  publisher = {PMLR}
}

@article{Ha25,
  author  = {Sehoon Ha and Joonho Lee and Michiel van de Panne and Zhaoming Xie and Wenhao Yu and Majid Khadiv},
  title   = {Learning-based Legged Locomotion: State of the Art and Future Perspectives},
  journal = {International Journal of Robotics Research},
  volume  = {44},
  number  = {8},
  pages   = {1396--1427},
  year    = {2025},
  doi     = {10.1177/02783649241312698}
}

@article{forcevla,
  author  = {Yu, Jiawen and Liu, Hairuo and Yu, Qiaojun and Ren, Jieji and Hao, Ce and Ding, Haitong and Huang, Guangyu and Huang, Guofan and Song, Yan and Cai, Panpan and Lu, Cewu and Zhang, Wenqiang},
  title   = {ForceVLA: Enhancing {VLA} Models with a Force-aware {MoE} for Contact-rich Manipulation},
  journal = {arXiv preprint arXiv:2505.22159},
  year    = {2025},
  note    = {NeurIPS 2025},
  doi     = {10.48550/arXiv.2505.22159},
  url     = {https://arxiv.org/abs/2505.22159}
}

@article{tactilevla,
  title         = {Tactile-{VLA}: Unlocking Vision-Language-Action Model's Physical Knowledge for Tactile Generalization},
  author        = {Huang, Jialei and Wang, Shuo and Lin, Fanqi and Hu, Yihang and Wen, Chuan and Gao, Yang},
  journal       = {CoRR},
  volume        = {abs/2507.09160},
  year          = {2025},
  doi           = {10.48550/arXiv.2507.09160},
  url           = {https://arxiv.org/abs/2507.09160},
  eprint        = {2507.09160},
  archivePrefix = {arXiv}
}

@article{sousa2014,
  title={Physical feasibility of robot base inertial parameter identification: A linear matrix inequality approach},
  author={Sousa, Crist{\'o}vao D and Cortesao, Rui},
  journal={The International Journal of Robotics Research},
  volume={33},
  number={6},
  pages={931--944},
  year={2014},
  publisher={SAGE Publications Sage UK: London, England}
}

@article{sindy,
author = {Steven L. Brunton  and Joshua L. Proctor  and J. Nathan Kutz },
title = {Discovering governing equations from data by sparse identification of nonlinear dynamical systems},
journal = {Proceedings of the National Academy of Sciences},
volume = {113},
number = {15},
pages = {3932-3937},
year = {2016},
doi = {10.1073/pnas.1517384113},
URL = {https://www.pnas.org/doi/abs/10.1073/pnas.1517384113},
eprint = {https://www.pnas.org/doi/pdf/10.1073/pnas.1517384113},
}

@article{pysindy, doi = {10.21105/joss.03994}, url = {https://doi.org/10.21105/joss.03994}, year = {2022}, publisher = {The Open Journal}, volume = {7}, number = {69}, pages = {3994}, author = {Kaptanoglu, Alan A. and de Silva, Brian M. and Fasel, Urban and Kaheman, Kadierdan and Goldschmidt, Andy J. and Callaham, Jared and Delahunt, Charles B. and Nicolaou, Zachary G. and Champion, Kathleen and Loiseau, Jean-Christophe and Kutz, J. Nathan and Brunton, Steven L.}, title = {PySINDy: A comprehensive Python package for robust sparse system identification}, journal = {Journal of Open Source Software} }

@article{pysr,
  title={Interpretable machine learning for science with PySR and SymbolicRegression. jl},
  author={Cranmer, Miles},
  journal={arXiv preprint arXiv:2305.01582},
  year={2023}
}

@article{Liu21,
  title={Sensorless force estimation for industrial robots using disturbance observer and neural learning of friction approximation},
  author={Liu, Sichao and Wang, Lihui and Wang, Xi Vincent},
  journal={Robotics and Computer-Integrated Manufacturing},
  volume={71},
  pages={102168},
  year={2021},
  publisher={Elsevier}
}

@ARTICLE{Iskandar23,
  author={Iskandar, Maged and Ott, Christian and Albu-Schäffer, Alin and Siciliano, Bruno and Dietrich, Alexander},
  journal={IEEE Robotics and Automation Letters}, 
  title={Hybrid Force-Impedance Control for Fast End-Effector Motions}, 
  year={2023},
  volume={8},
  number={7},
  pages={3931-3938},
  doi={10.1109/LRA.2023.3270036}}

@ARTICLE{Carron,
  author={Carron, Andrea and Arcari, Elena and Wermelinger, Martin and Hewing, Lukas and Hutter, Marco and Zeilinger, Melanie N.},
  journal={IEEE Robotics and Automation Letters}, 
  title={Data-Driven Model Predictive Control for Trajectory Tracking With a Robotic Arm}, 
  year={2019},
  volume={4},
  number={4},
  pages={3758-3765},
  doi={10.1109/LRA.2019.2929987}}

@article{atkeson,
  title={Estimation of Inertial Parameters of Manipulator Loads and Links},
  author={Atkeson, Christopher G and An, Chae H and Hollerbach, John M},
  journal={The International Journal of Robotics Research},
  volume={5},
  number={3},
  pages={101--119},
  year={1986},
  publisher={SAGE Publications}
}



\end{document}